\documentclass[sigconf]{acmart}

\usepackage{booktabs} % For formal tables
\usepackage{tikz, pgfplots}
\usepackage{subfig}
\usepackage[para]{footmisc}
\usepackage{multirow}

\setcopyright{rightsretained}
\renewcommand\footnotetextcopyrightpermission[1]{} % removes footnote with conference information in first column
\acmDOI{10.475/123_4}

\acmISBN{123-4567-24-567/08/06}

\acmConference[]{}{}{}
\acmYear{2018}
\copyrightyear{2018}

\acmArticle{4}
\acmPrice{15.00}

\begin{document}
\title[Research Paper]{Instance-based Inductive Deep Transfer Learning by Cross-Dataset Querying with Locality Sensitive Hashing}
%\titlenote{Produces the permission block, and copyright information}
%\subtitle{Extended Abstract}
%\subtitlenote{The full version of the author's guide is available as \texttt{acmart.pdf} document}

\author{Somnath Basu Roy Chowdhury}
%\authornote{Dr.~Trovato insisted his name be first.}
\affiliation{
  \institution{IIT Kharagpur}
  %\streetaddress{P.O. Box 1212}
}
\email{brcsomnath@ee.iitkgp.ernet.in}

\author{Annervaz K M}
\affiliation{
	\institution{Indian Institute of Science \& Accenture Technology Labs}
}
\email{annervaz@iisc.ac.in}

\author{Ambedkar Dukkipati}
\affiliation{
	\institution{Indian Institute of Science}
}
\email{ambedkar@iisc.ac.in}


\begin{abstract}
Supervised learning models are typically trained on a single dataset and the performance of these models rely heavily on the size of the dataset, i.e., amount of data available with the ground truth. Learning algorithms try to generalize solely based on the data that is presented with during the training. In this work, we propose an inductive transfer learning method that can augment learning models by infusing similar instances from different learning tasks in the Natural Language Processing (NLP) domain. We propose to use instance representations from a source dataset, \textit{without inheriting anything} from the source learning model. Representations of the instances of \textit{source} \& \textit{target} datasets are learned, retrieval of relevant source instances is performed using soft-attention mechanism and \textit{locality sensitive hashing}, and then, augmented into the model during training on the target dataset. Our approach simultaneously exploits the local \textit{instance level information} as well as the macro statistical viewpoint of the dataset. Using this approach we have shown significant improvements for three major news classification datasets over the baseline. Experimental evaluations also show that the proposed approach reduces dependency on labeled data by a significant margin for comparable performance. With our proposed cross dataset learning procedure we show that one can achieve competitive/better performance than learning from a single dataset. 
\end{abstract}

%\begin{CCSXML}
%<ccs2012>
 % <concept_id>10010520.10010553.10010562</concept_id>
 % <concept_desc>Computer systems organization~Embedded systems</concept_desc>
 %</concept>
 %<concept>
 % <concept_id>10010520.10010575.10010755</concept_id>
 % <concept_desc>Computer systems organization~Redundancy</concept_desc>
 % <concept_significance>300</concept_significance>
 %</concept>
 %<concept>
 % <concept_id>10010520.10010553.10010554</concept_id>
 % <concept_desc>Computer systems organization~Robotics</concept_desc>
 % <concept_significance>100</concept_significance>
 %</concept>
 % <concept_id>10003033.10003083.10003095</concept_id>
  %<concept_desc>Networks~Network reliability</concept_desc>
  %<concept_significance>100</concept_significance>
 %</concept>
%</ccs2012>
%\end{CCSXML}
\maketitle

\section{Introduction \& Motivation}
A fundamental issue with supervised learning techniques (like classification) is the requirement of enormous amount of labeled data, which in some scenarios maybe expensive to gather or may not be available. Every supervised task requires a separate labeled dataset and training state-of-the-art deep learning models is computationally expensive for large datasets. In this paper, we propose a deep transfer learning method that can enhance the performance of learning models by incorporating information from a different dataset, encoded while training for a different task in a similar domain. 

The approaches like transfer learning and domain adaptation have been studied extensively to improve adaptation of learning models across different tasks or datasets. In transfer learning, certain portions of the learning model are re-trained for fine-tuning weights in order to fit a subset of the original learning task. Transfer learning suffers heavily from domain inconsistency between tasks and may even have a negative effect~\cite{rosenstein2005transfer} on performance. Domain adaptation techniques aim to predict unlabeled data given a pool of labeled data from a similar domain. In domain adaptation, the aim is to have better generalization as source and target instances are assumed to be coming from different probability distributions, even when the underlying task is same.

We present our approach in an \textit{inductive transfer learning}~\cite{pan2010survey} framework, with a labeled \textit{source} (domain $\mathcal{D}_S$ and task $\mathcal{T}_S$) and \textit{target} (domain $\mathcal{D}_T$ and task $\mathcal{T}_T$) dataset, the aim is to boost the performance of target predictive function $f_T(\cdot)$ using available knowledge in $\mathcal{D}_S$ and $\mathcal{T}_S$, given $\mathcal{T}_S \neq \mathcal{T}_T$. We retrieve instances from $\mathcal{D}_S$ based on similarity criteria with instances from $\mathcal{D}_T$, and use these instances while training to learn the target predictive function $f_T(\cdot)$. We utilize the instance-level information in the source dataset, and also make the newly learnt target instance representation similar to the retrieved source instances. This allows the learning algorithm to improve generalization across the source and target datasets. We use \textit{instance-based learning} that actively looks for similar instances in the source dataset given a target instance. The intuition behind retrieving similar instances comes from an instance-based learning perspective, where simplification of the class distribution takes place within the locality of a test instance. As a result, modeling of similar instances become easier~\cite{aggarwal2014instance}. Similar instances have the maximum amount of information necessary to classify an unseen instance, as exploited by techniques like $k$-nearest neighbours.

We derived inspiration for this method from the working of the human brain, where new memory representations are consolidated, slowly over time for efficient retrieval in future. According to~\cite{mcgaugh2000memory}, newly learnt memory representations remain in a fragile state and are affected as further learning takes place. In our procedure, we make use of encodings of instances precipitated while training for a different task using a different model. This being used for a totally different task, and adapted as needed, is in alignment with \textit{memory consolidation} in human brain. 

An attractive feature of the proposed method is that the search mechanism allows us to use more than one source dataset during training to achieve inductive transfer learning. Our approach differs from the standard instance-based learning in two major aspects. First, the instances retrieved are not necessarily from the same dataset, but can be from various secondary datasets. Secondly, our model simultaneously makes use of local instance level information as well as the macro-statistical viewpoint of the dataset, where typical lazy instance based learning like $k$-nearest neighbour search make use of only the local instance level information. In order to ensure that the learnt latent representations can be utilized by another task, we try to make the representations similar. The need for this arise as we need to ensure that similar instances in two different domain have similar representations.

\textbf{Motivating Example}. BBC\footnote{http://www.bbc.com/} and
SkySports\footnote{http://www.skysports.com/}, two popular news
channels are used to illustrate the example. BBC reports news about
all domains in daily life, on the other hand SkySports focuses only on
sports news. However if BBC decides to restructure its sports section
depending on the type of sport, we need to have a supervised
classifier to achieve this goal. BBC although has a significant amount
of sports news article, it lacks significant amount of labeled sports news articles in order
to build a reliable classifier. Instance-based learning techniques
will not perform well in such a situation. The ability of the proposed method
to give competitive performance with limited training data, by making use
of labeled training data from existing dataset helps in the scenario. 
Labeled data from SkySports can be incorporated to achieve
this goal of classifying news articles. Similarly this approach can be
extended to gather instances from multiple news channels other than
SkySports to enhance the performance of such a classifier, with labeling
fewer samples from BBC.

We develop our instance retrieval based transfer
learning technique, which is capable of extracting information from
multiple datasets simultaneously in order to tackle the problem of
limited labeled data or unbalanced labeled dataset. We also enforce
constraints to ensure the learning model learns representations
similar to the external source domains, thereby aiding in the
classification model. To the best of our knowledge this is the first
work which unifies instance-based learning in transfer learning
setting.

The main contribution of the work are as follows,
\begin{enumerate}
	\item We propose an augmented neural network model for combining instance and model based learning.
	\item We use Locality Sensitive Hashing for effective retrieval of similar instances efficiently in sub-linear time and fuse it to the learning model.
	\item We hypothesize and illustrate with detailed experimental results, performance of the learning models can be improved by infusing instance level information from within the dataset and across datasets. In both these experiments we show an improvement of  5+\% over the baseline.
	\item Proposed approach is shown to be useful for training on very lean datasets, by leveraging support from large datasets.
\end{enumerate}

\section{Background}
For instance transfer to take place in a deep learning framework, natural language sentences are converted into a vector representation in a latent space. Long Short-Term Memory (LSTM) networks with randomly initialized word embeddings act as our baseline model. Once the sentences are encoded in their numerical representations we apply similarity search across source dataset instances using Locality Sensitive Hashing (LSH). In this section, we briefly summarize LSH and transfer learning to clarify the setup of our work, in an inductive transfer learning setting.

%\subsection*{Long Short Term Memory (LSTM)}
%Long short-term memory (LSTM)~\cite{hochreiter1997long} is the most popular recurrent neural network (RNN) architecture that deals with the vanishing gradient problem~\cite{pascanu2013difficulty} by introduction of supplementary gates. RNN architecture takes up the general form $$h_t = f(h_{t-1}, x_t, \theta)$$ where $h_t$ and $ x_t$ are the hidden state and input at time $t$ respectively, $\theta$ the model parameters. The LSTM variant introduces 3 additional gates namely input, output and forget gates. The gates in the LSTM architecture is governed using the following equations $$f_t = \sigma(W_{fh}h_{t-1} + W_{fx}xt + b_f )$$
%$$i_t = \sigma(W_{ih}h_{t-1} + W_{ix}x_t + b_i)$$
%$$o_t = \sigma(W_{oh}h_{t-1} + W_{ox}x_t+b_o)$$
%$$\tilde{c}_t = tanh(W_{ch}h_{t-1} + W_{cx}x_t + b_c)$$
%$$c_t = f_t\odot c_{t-1} + i_t\odot \tilde{c}t$$
%$$h_t = o_t \odot \tanh(c_t)$$
%where $\sigma(\cdot)$ denotes the element-wise sigmoid function, $W_{\_\_}$'s, $b_\_$'s denote model parameters and  $\odot$ denotes element-wise multiplication. LSTM has the capability to modify and retain the content on the cell state as  required by the task, using the gates and hidden states. While a forward LSTM takes the input sequence as it is, a backward LSTM takes the input in the reverse order. A backward LSTM is used to capture the dependencies of a word on future words in the original sequence. A concatenation of a forward LSTM and a backward LSTM is known as bi-directional LSTM (bi-LSTM)~\citep{greff2015lstm}.

\begin{figure*}
	\includegraphics[width=\textwidth, keepaspectratio]{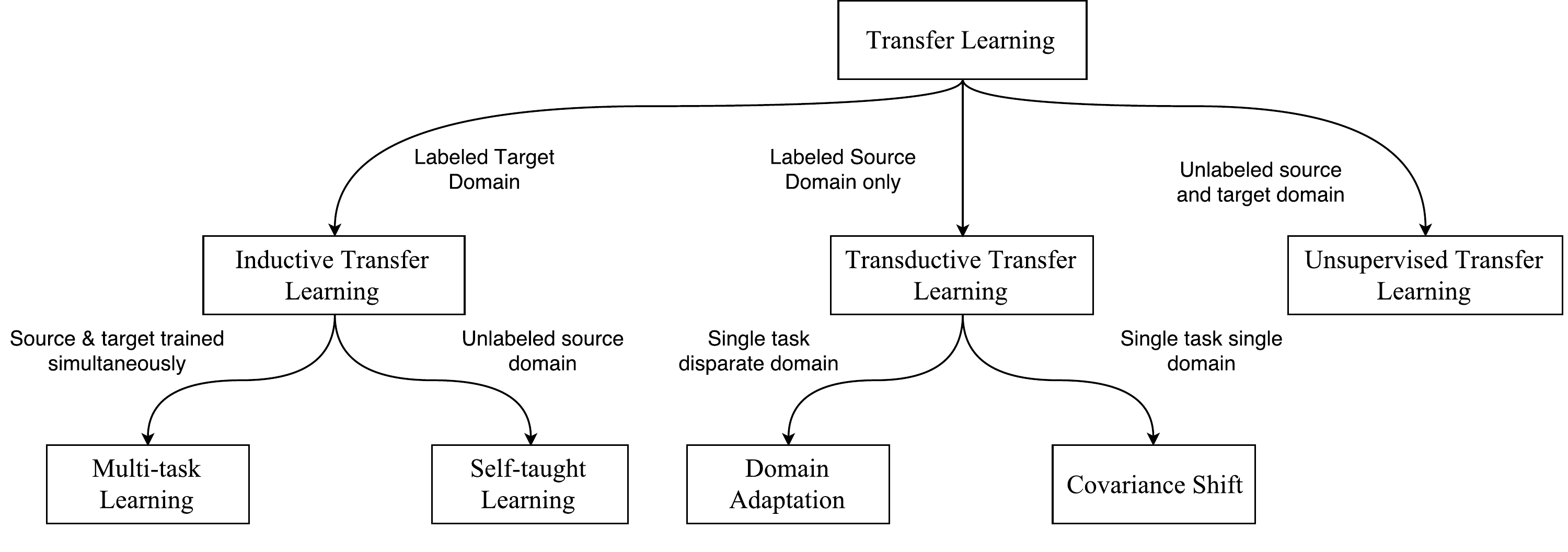}
	\caption{Variations in Transfer Learning settings}
	\label{fig:TL}
\end{figure*}

\subsection{Locality Sensitive Hashing (LSH)}
Locality Sensitive Hashing
\cite{gao2014dsh,gionis1999similarity}
is an algorithm which performs approximate nearest neighbor similarity
search for high-dimensional data in sub-linear time. The main
intuition behind this algorithm is to form LSH index for each point
which maps "similar" points to the same bucket with higher
probability. Approximate nearest neighbors of a query is retrieved by
hashing it to a bucket and returning other points from the
corresponding bucket. 

The locality sensitive hash family, $\mathcal{H}$ has to satisfy
certain constraints mentioned in~\cite{indyk1998approximate} for
nearest neighbor retrieval. The LSH Index maps each point $p$ into a
bucket in a hash table with a label $g(p) = (h_1(p), h_2(p), \ldots,
h_k(p))$, where $h_1, h_2, \ldots, h_k$ are chosen independently with
replacement from $\mathcal{H}$. We generate $l$ different hash
functions of length $k$ given by $G_j(p) = (h_{1j}(p), h_{2j}(p),
\cdots,$ $ h_{kj}(p))$ where $j \in {1, 2, \ldots, l}$ denotes the
index of the hash table. Given a collection of data points
$\mathcal{C}$, we hash them into $l$ hash tables by concatenating
randomly sampled $k$ hash functions from $\mathcal{H}$ for each hash
table. While returning the nearest neighbors of a query Q, it is
mapped into a bucket in each of the $l$ hash tables. The union of all
points in the buckets $G_j(Q), j = {1, 2, \ldots, l}$ is
returned. Therefore, all points in the collection $\mathcal{C}$ is not
scanned and the query is executed in sub-linear time. The storage
overhead for LSH is sub-quadratic in $n$, the number of points in the
collection $\mathcal{C}$. 

LSH Forests~\cite{bawa2005lsh} are an improvement over LSH Index which
relaxes the constraints on hash family $\mathcal{H}$ with better
practical performance guarantees. LSH Forests utilizes $l$ prefix
trees (LSH trees) instead of having hash tables, each constructed from
independently drawn hash functions from $\mathcal{H}$. The hash
function of each prefix tree is of variable length ($k$) with an upper
bound $k_m$. The length of the hash label of a point is increased
whenever a collision occurs to form leaf nodes from the parent node in
the LSH tree. For $m$ nearest neighbour query of a point $p$, the $l$
prefix trees are traversed in a top-down manner to find the leaf node
with highest similarity with point $p$. From the leaf node, we
traverse in a bottom-up fashion to collect $M$ points from the forest,
where $M = cl$, $c$ being a small constant. It has been shown in
~\cite{bawa2005lsh}, that for practical cases the LSH Forests execute
each query in constant time with storage cost linear in $n$, the
number of points in the collection $\mathcal{C}$. 

\subsection{Transfer Learning}
Traditional machine learning algorithms try to learn a statistical
model which is capable of predicting unseen data points, given that it
has been trained on labeled or unlabeled training samples. In order to
reduce the dependency on data, the need to reuse knowledge across
tasks arise. \textit{Transfer learning} allows such knowledge transfer
to take place even if the domain, tasks and distribution of the
datasets are different. %Given a source domain $\mathcal{D}_S$ and corresponding task $\mathcal{T}_S$, and a target domain $\mathcal{D}_T$ and corresponding task $\mathcal{T}_T$, the goal of transfer learning is to enhance the learning process of the target function $f_T(\cdot)$, using information present in $\mathcal{D}_S$ and $\mathcal{T}_S$, where $\mathcal{D}_S \neq \mathcal{D}_T$ and $\mathcal{T}_S \neq \mathcal{T}_T$.  
Transfer learning can be applied in various problem frameworks,
depending on the nature of source and target domain. Based on these
variations, it can be broadly classified
into three  categories (a) \textit{inductive transfer learning}
(b) \textit{transductive transfer learning} and (c)
\textit{unsupervised transfer learning}. Figure~\ref{fig:TL} shows the various problem settings and its corresponding transfer learning setup.  We will discuss the fundamental differences in the operation of these methods here. 

\textbf{Inductive transfer learning}. In this setup, labeled data is
available in the target domain to \textit{induce} the prediction
function in target domain $\mathcal{D}_T$. The target and source tasks
are different $\mathcal{T}_S \neq \mathcal{T}_T$, however they may or
may not share a common domain. Inductive transfer learning can be further classified into two sub-categories where (a) labeled source
instances are available and where (b) ground-truth for source instances
are absent (\textit{self-taught learning}~\cite{raina2007self}). 

\textbf{Transductive transfer learning}. In this setting the source
and target tasks are same $\mathcal{T}_S = \mathcal{T}_T$, while their
domains are different $\mathcal{D}_S \neq
\mathcal{D}_T$. This technique is also sub-divided into two categories
where (a) the learning algorithm considers source and target domain to
be different and have a separate feature space and where (b) the
feature space is same in an attempt to reduce domain discrepancy, this
is also known as \textit{domain adaptation}~\cite{daume2006domain}. 

\textbf{Unsupervised transfer learning}. In this framework, the source
and domain tasks are related but different $\mathcal{T}_S \neq
\mathcal{T}_T$. Both source and target domains have unlabeled
instances, this techniques is used in unsupervised task settings like
dimensionality reduction~\cite{wang2008transferred}, cluster
approximation~\cite{dai2008self} etc. 

 In this paper, our contribution is presented in \textit{inductive transfer
   learning} framework. Knowledge transfer in this setup takes place
 in four ways (a) instance-transfer %~\cite{dai2007boosting,    zadrozny2004learning, dai2007transferring, quionero2009dataset,   jiang2007instance} 
 (b) feature-representation-transfer (c)
 parameter-transfer and (d) relational-knowledge-transfer. Parameter transfer and  relational-knowledge transfer are studied exhaustively in inductive  transfer literature. 
 In our proposed approach we infuse instance-level
 feature representation transfer across source and target domain, in
 order to enhance the learning process.
\begin{figure*}
	\includegraphics[width=\textwidth, keepaspectratio]{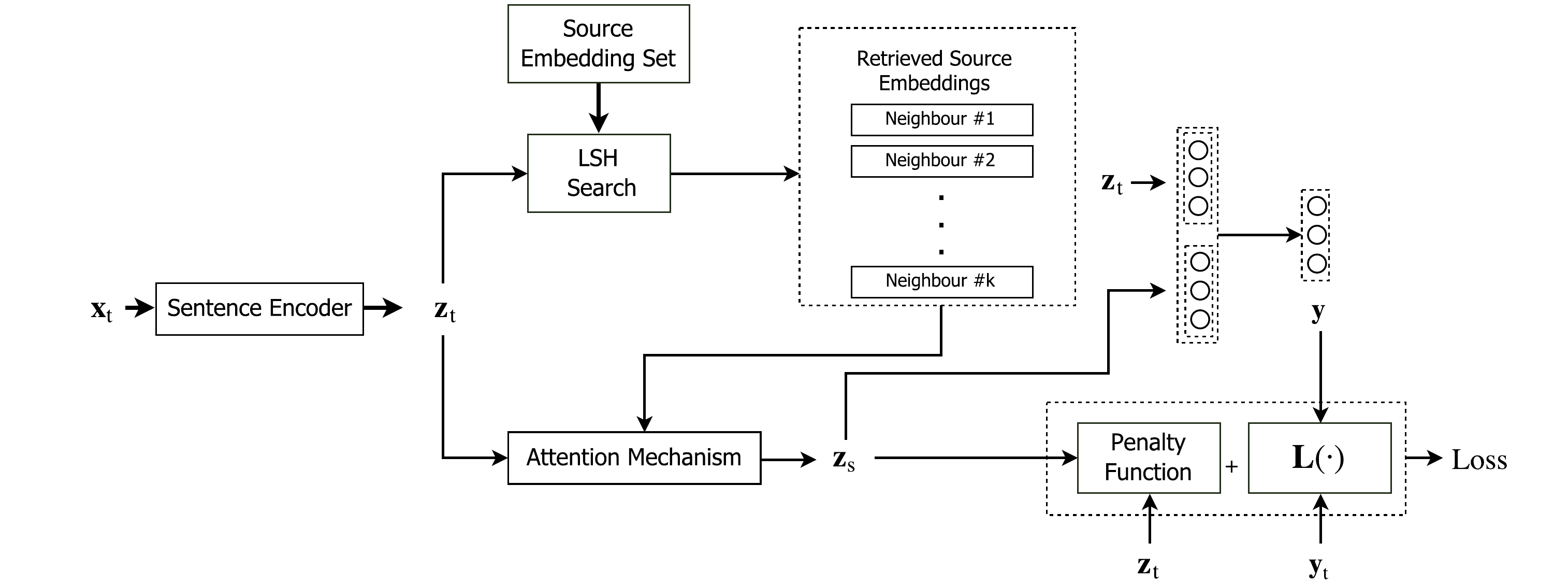}
	\caption{Proposed Model Architecture}
	\label{fig:approach}
\end{figure*}

\section{Proposed Model}
%In this section, we describe our model and the details of its building
%blocks. Before getting into the model details, let us see
%mathematically what we are trying to achieve.
Given the data $x$ with the ground truth $y$, supervised learning models
aim at finding the parameters $\Theta$ that maximizes the log-likelihood
as
\begin{displaymath}
\Theta = \mathop{\mathrm{arg max}}\limits_{\Theta} \log P(y|
\mathrm{x}, \Theta).
\end{displaymath}
We propose to augment the learning by infusing  similar instances latent
representations $\mathrm{z}_s$, from a source dataset, a latent vector
from source  dataset $\mathrm{z}_s$ is retrieved using the data
sample $\mathrm{x}_t$ (target dataset instance). Thus, our modified
objective function  can be expressed as
\begin{displaymath}
\mathop{\mathrm{max}}\limits_{\Theta} P(y| \mathrm{x}_t, \mathrm{z}_s,
\Theta).
\end{displaymath}
To enforce latent representations of the instances to be similar, for better 
generalization across the tasks, we add a suitable penalty to the objective.
The modified objective then becomes,
\begin{displaymath}
\Theta = \mathop{\mathrm{arg max}}\limits_{\Theta}  \log{P}(y|
\mathrm{x}_t, \mathrm{x}_s, \Theta) - \lambda\mathcal{L}(\mathrm{z}_s,
\mathrm{z}_t),
\end{displaymath}
where $\mathcal{L}$ is the penalty function and $\lambda$ is a hyperparmeter. 

%This is schematically represented in figure~\ref{fig:approach}
The subsequent sections focus on the methods to retrieve instance latent vector $\mathrm{z}_s$ using the data sample $\mathrm{x}_t$. It is important to note that, we do not assume any structural form for $P$. Hence the proposed method is applicable to augment any supervised learning setting with any form for $P$. In the experiments we have used softmax~\cite{bishop2006pattern} using the bi-LSTM ~\cite{greff2015lstm} encodings of the input as the form for $P$. The schematic representation of the model is shown in Figure~\ref{fig:approach}. In the following sections, we will discuss the in-detail working of individual modules in Figure~\ref{fig:approach} and formulation of the penalty function $\mathcal{L}$ .

\subsection{Sentence Encoder}
\label{encoder}
The purpose of this module is to create a vector in a latent space by encoding the 
semantic context of a sentence from the input sequence of words. The
context vector $c$  is obtained from an input sentence which is a
sequence of word vectors $\mathbf{x} = (x_1, x_2, \ldots, x_T)$, using
a bi-LSTM (Sentence Encoder shown in Figure~\ref{fig:approach}) as 
\begin{displaymath}
h_t = f(x_t, h_{t-1}),
\end{displaymath}
where $h_t \in \mathbb{R}^n$ is the hidden state of the bi-LSTM  at time
$t$ and $n$ is the embedding size.
We combine the states at multiple time steps using a linear function g. We have, 
\begin{displaymath}
o = g(\{h_1, h_2, \ldots, h_T\})\:\: \mbox{and} \:\:\:c = \mathrm{ReLU}(o^TW),
\end{displaymath}
where $W \in \mathbb{R}^{n \times m}$ and $m$ is a hyper parameter
representing the dimension of the context vector. $g$ in our experiments is set as
\begin{displaymath}
g(\{h_1, h_2, \ldots, h_T\}) = \frac{1}{T}{\sum_{t=1}^{T}h_t}.
\end{displaymath}  
The bi-LSTM  module responsible for generating the context vector $c$ is pre-trained on the target classification task. A separate bi-LSTM module (sentence encoder for the source dataset) is trained on the source classification task to obtain instance embeddings for the target dataset. In our experiments we used similar modules for creating the instance embeddings of the source and target dataset, this is not constrained by the method and different modules can be used here.

\subsection{Instance Retrieval}
\label{retrieval}
%In this section, we will discuss the LSH Search and Attention
%Mechanism in the schematic diagram shown in
%Figure~\ref{fig:approach}.
Using the obtained context vector $c_t$ ($c$ in Section~\ref{encoder})
corresponding to a target instance as a query, $k$-nearest neighbours
are searched from the source dataset $(z_1^s, z_2^s, \ldots, z_k^s)$
using LSH. The search mechanism using LSH takes constant time in
practical scenarios~\cite{bawa2005lsh} and therefore doesn't affect the
training duration by large margins. The retrieved source dataset
instance embeddings receive attention $\alpha_i^z$, using
soft-attention mechanism based on cosine similarity given as, 
\begin{displaymath}
\alpha_{i}^z = \frac{\exp(c^T_tz_i^s)}{\sum\limits_{j=1}^{k}
  \exp(c^T_tz_j^s)}, %\hspace*{0.5cm} \sum_{i=1}^{k} \alpha_{i}^z =
                     %1$$
\end{displaymath}
where $c \in \mathbb{R}^{m}$ and $z_i^s, z_j^s \in \mathbb{R}^{m}$.

The fused instance embedding vector $z_s$ formed after soft attention
mechanism is given by,
\begin{displaymath}
z_s = \sum_{i=1}^{k} \alpha_{i}^z z_i^s,
\end{displaymath}
where $z_s \in \mathbb{R}^{m}$. The retrieved instance is concatenated
with the context vector $c$ from the classification module as
\begin{displaymath}
\mathrm{s} = [c_t, z_s]\:\: \mbox{and}\:\: \mathbf{y} = \mathrm{softmax}(\mathrm{s}^TW^{(1)}),
\end{displaymath}
where $W^{(1)} \in \mathbb{R}^{2m \times u}$, $\mathbf{y}$ is the output of the final target classification task. 
This model is then trained jointly with the initial parameters from the pre-trained classification module. The pre-training of the classification module is necessary because if we start from a randomly initialized context vector $c_t$, the LSH Forest retrieves arbitrary vectors and the model as a whole fails to converge. As the gradient only propagates through the
attention values and penalty function it is impossible to simultaneously rectify the query and search results of the hashing
mechanism.  

It is important to note that the proposed model adds only a limited
number of parameters over the baseline model. The extra trainable weight matrix in the
model is $W^{(1)} \in \mathbb{R}^{2m \times u}$, adding only
$2m \times u$, where $m$ is the size of the context vector $c$ and $u$
is the number of classes.
%The model achieves significant improvement
%across all datasets by infusing local instance level information
%from a supporting dataset using only 
%a limited number of extra parameters compared to the baseline model. 

\subsection{Instance Clustering}
\label{clustering}
While training our model, instances are retrieved in an online manner using LSH. In the case of large source datasets, where
the number of instances is in the range of millions, the LSH becomes really slow and training may take impractical amount of time. In order to overcome this problem, the source instances are clustered and the centroid of the clusters formed are considered as our search entities.  

\begin{figure}[h!]
	\centering
\subfloat[Original latent vector space]{\includegraphics[scale=0.2]{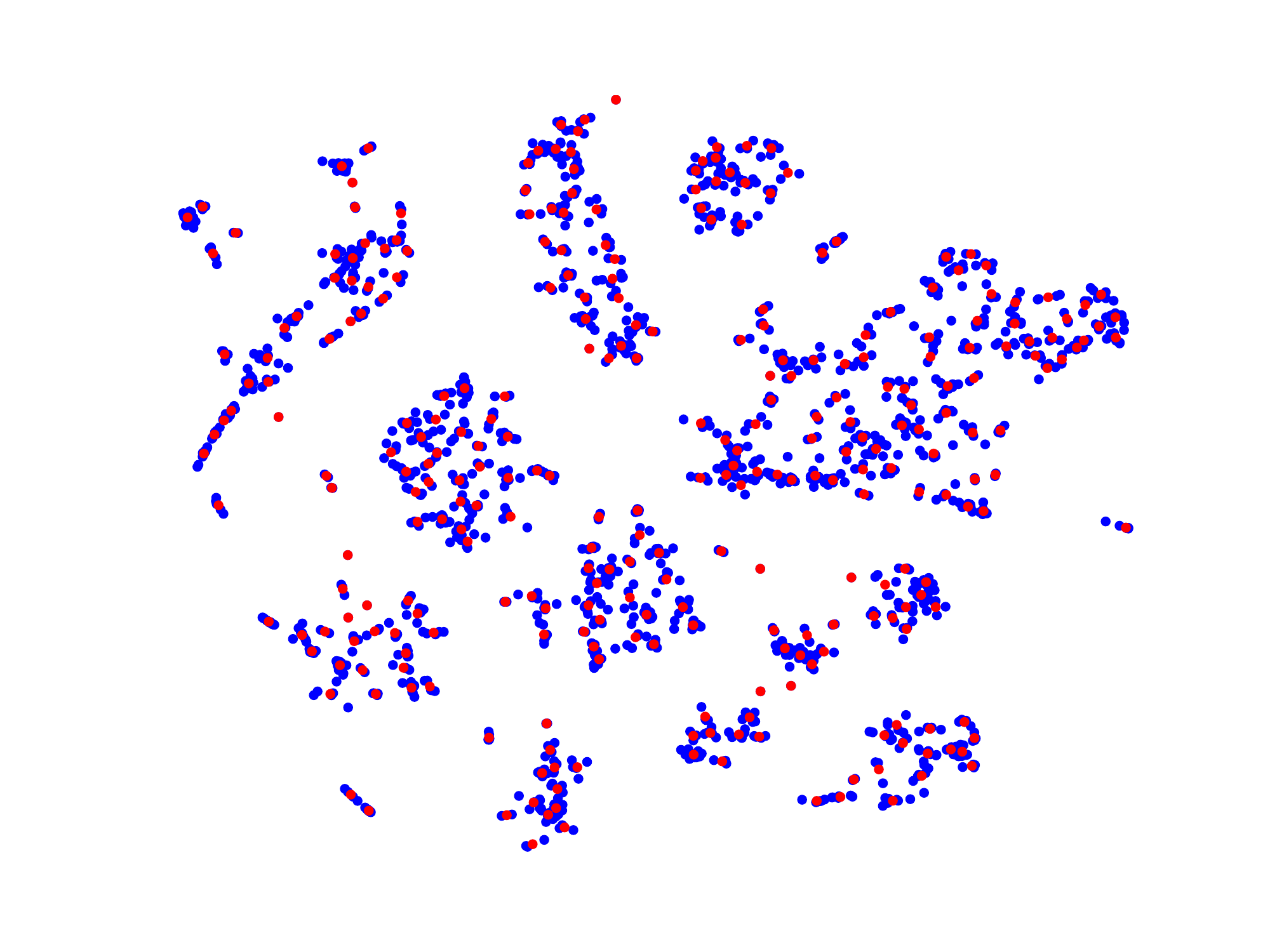}}
\subfloat[Clustered vector space]{\includegraphics[scale=0.2]{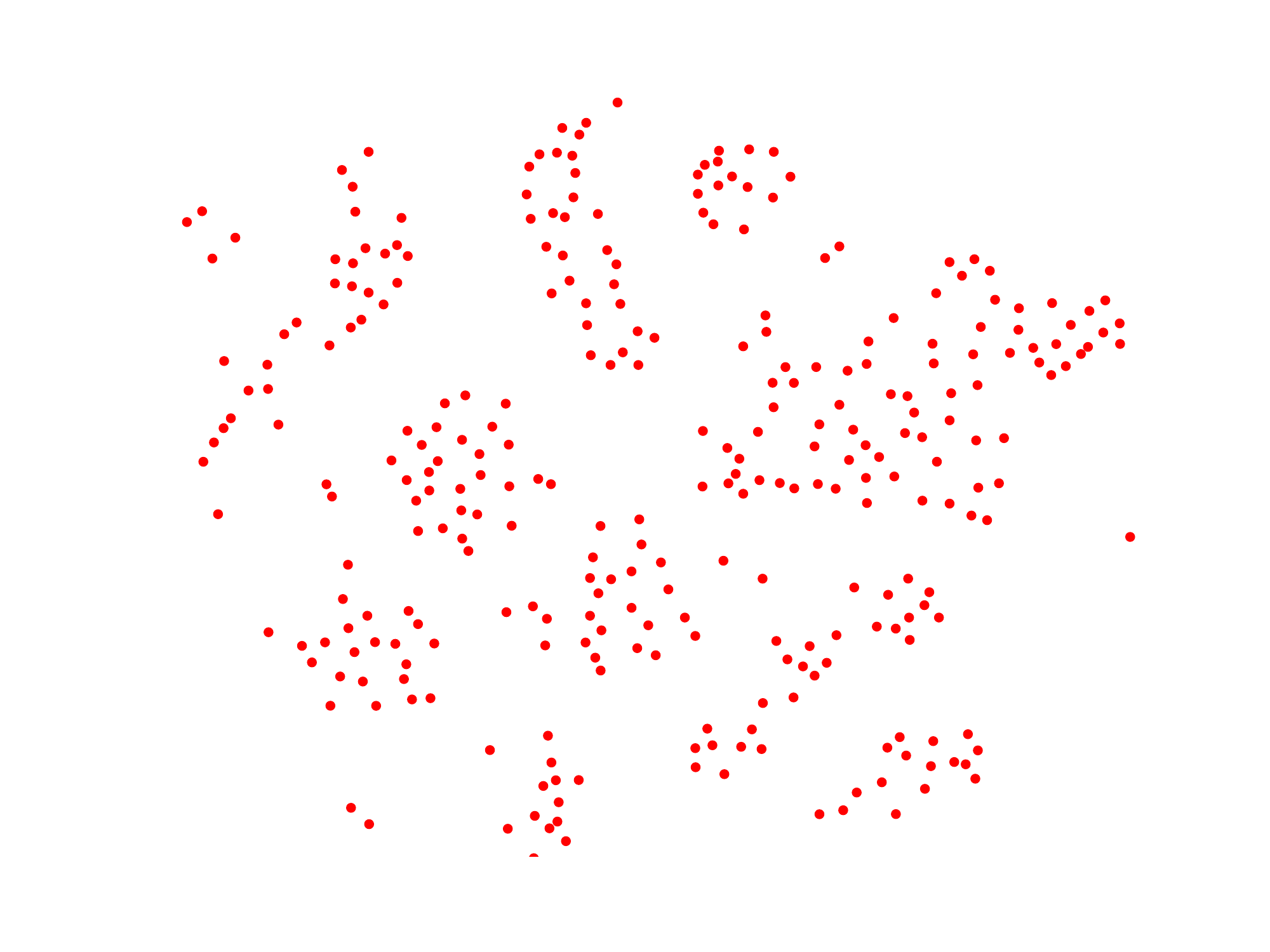}}
	\caption{The figure shows t-SNE visualizations of latent vectors obtained using bi-LSTM module for BBC dataset (a) original vectors with cluster centers marked in red (b) sparse latent vector space obtained using k-means clustering.}
	\label{cluster}
\end{figure}

Fast $k$-means clustering~\cite{shindler2011fast} is used in the
clustering process as the number of instances and clusters are quite
large in this setup. The number of clusters is set to an upper limit
of 10000, as LSH search performance is significantly fast with this
search space. Figure~\ref{cluster} shows the
t-SNE~\cite{maaten2008visualizing} visualization for BBC
dataset. Figure~\ref{cluster} (a) shows the latent vector space of the
entire dataset with the cluster centers marked in red,
Figure~\ref{cluster} (b) shows the cluster centers forming a sparse
representation of the latent vector embeddings which are used in the
experiment for classification. 

\subsection{Penalty Function}
In an instance-based learning, a test instance is assigned the label of
the majority of its nearest-neighbour instances. This follows from the
fact that similar instances belong to the same class
distribution. Following the retrieval of latent vector embeddings from
the source dataset, the target latent embedding is constrained to be
similar to the retrieved source instances. In order to enforce this,
we introduce an additional penalty along with the loss function (shown
in Figure~\ref{fig:approach}). The modified objective function is
given as
\begin{displaymath}
\min_{\theta} L(\mathbf{y}, y_t) + \lambda||z_s - z_t||_2^2\enspace,
\end{displaymath}
where $\mathbf{y}$ and $z_s$ are the outputs of the model and retrieved
latent embedding respectively (as in Section~\ref{retrieval}), $y_t$ is
the label, $\lambda$ is scaling factor and $z_t$ is the latent vector
embedding of the target instance. $L(\cdot)$ in the above
equation denotes the loss function used to train the model (depicted as \textbf{L($\cdot$)} in Figure~\ref{fig:approach}) and $\theta$ denotes the model
parameters.  The additional penalty term  enables the latent vectors to be similar across multiple datasets, which aids performance in the subsequent stages.

\section{Experiments}
\label{exp}
The experiments are designed in a manner to compare the performance of the baseline model with that of external dataset augmented model. Our experiments shows performance enhancement across several datasets by incorporating relevant instance information from a  source dataset in varying setups.  Our experiments also illustrate that our proposed model continues to perform better even when the size of training set is reduced, thereby reducing the dependence on labeled data.  We also demonstrate the efficacy of our model through latent vector visualizations.

\textbf{Baseline}. A simple \textit{bi-LSTM (target-only)} model is trained without consideration for source-domain instances (no source-instance retrieval branch included into the network), this is used as the baseline. The \textit{Instance-infused bi-LSTM} model is trained on the target domain with class labels revealed. This model serves as a tool to gain available knowledge and consolidate available representations in light of the past knowledge, assuming that source data is relevant for the downstream task at hand.

 %The embeddings of the source instances are also trained using bi-LSTM classifier. The only constraint on the embeddings is that their shape should be same across multiple domain for \textit{locality sensitive hashing} search to take place. As there is no bound on the entries in the latent vector, the LSH search takes place in a space where the vector entries are passed through a \textit{softmax} module for uniformity. As the instance queries are processed in an offline manner (no learning involved in it), the \textit{bi-lstm (target only)} model is pre-trained so as to proper query initialization for LSH. In absence of proper query initialization, with randomly initialized queries similar instances are retrieved and the \textit{instance-infused} model fails to converge as a whole. In the following subsections we discuss the experimental setup, datasets and insights into the experiments performed. We will make our code public after the review.

\subsection{Datasets}
For our experiments, we have chosen three popular publicly-available news classification datasets. The datasets share common domain information and their details regarding the three popular datasets are mentioned here
\begin{enumerate}
	\item \textbf{20 Newsgroups (News20)}\footnote{http://qwone.com/~jason/20Newsgroups/}: A collection of news group articles in English~\cite{Lichman:2013}. The dataset is partitioned almost evenly across 20 different classes: \textit{comp.graphics}, \textit{comp.os.ms-windows.misc}, \textit{comp.sys.ibm.pc.hardware}, \textit{comp.sys.mac. hardware}, \textit{comp. windows.x}, \textit{rec.autos}, \textit{rec.motorcycles}, \textit{rec.sport. baseball}, \textit{rec.sport. hockey, sci.crypt, sci.electronics,	sci.med, sci.space, misc.forsale, talk.politics.misc, talk.politics.guns, talk. politics.mideast, talk. religion.misc, alt.atheism} and \textit{soc. religion.christian}.
	\item \textbf{BBC}\footnote{\label{bbc}http://mlg.ucd.ie/datasets/bbc.html}: Original news article from BBC (2004-2005) in English~\cite{greene06icml},  classified into 5 classes: \textit{business, entertainment, politics, sport} and \textit{tech}.
	\item \textbf{BBC Sports}\textsuperscript{\ref{bbc}}: Sports news articles from BBC news ~\cite{greene06icml}. The dataset is divided into 5 major classes: \textit{athletics, cricket, football, rugby} and \textit{tennis}.
\end{enumerate}
 The datasets are chosen in such a way that all of them share common domain knowledge and have small number of training examples so that the improvement observed using instance-infusion is significant. The statistics of the three real-world datasets are mentioned in Table ~\ref{dataset}.
  \begin{table}[h!]
  	\centering
  	\begin{tabular}{ c c c c}
  		\hline 
  		&&& \\[-1em]
  		\textsc{Dataset} & \textsc{Train Size} & \textsc{Test Size} & \# \textsc{Classes}\\
  		\hline
  		&&& \\[-1em]
  		\textsc{News20} & 18000 & 2000 & 20 \\ 
  		\textsc{BBC} & 2000 & 225 & 5\\ 
  		\textsc{BBC Sports} & 660 & 77 & 5\\ 
  		\hline
  	\end{tabular}\\
  	\vspace*{1mm}
  	\caption{Dataset Specifications}
  	\label{dataset}
  \end{table}
  
The mentioned datasets do not have a dedicated test set, so the evaluations were performed using $k$-\textit{fold cross validation} scheme. All performance scores that are reported in this paper are the mean performance over all the folds.

\begin{table}[h!]
	\centering
	\begin{tabular}{ c  c  c c}
		\hline\\[-1em]
		\textbf{Hyper-parameter} & \textsc{News20} & \textsc{BBC} & \textsc{BBC-Sports}\\
		\\[-1em]
		\hline
		\\[-1em]
		Batch size & 256 & 32 & 16\\ 
		Learning rate & 0.01 & 0.01 & 0.01\\
		Word vector dim & 300 & 300 & 300\\
		Latent vector dim ($m$) & 50 & 50 & 50\\
		\# Nearest neighbours ($k$) & 5 & 5 & 5\\
		Scaling factor ($\lambda$) & $10^{-4}$ & $10^{-4}$ & $10^{-4}$\\
		\# Epochs per fold & 30 & 20 & 20\\
		\hline
	\end{tabular}
	\vspace*{1mm}
	\caption{Hyper-parameters which were used in experiments for News20, BBC \& BBC-Sports datasets}
	\label{tab:hyperparameters}
\end{table}

 \begin{table*}[h!]
 	\centering
 	\begin{tabular}{ p{6.5cm} c| c c c c c c}
 		\hline \\[-0.5em]
 		& \textsc{Target} &\multicolumn{2}{c}{\textsc{News20}}  & \multicolumn{2}{c}{\textsc{BBC}}& \multicolumn{2}{c}{\textsc{BBC Sports}}\\
 		\multirow{2}{*}{\textbf{\textsc{METHOD}}} & \textsc{Source} &  \multicolumn{2}{c}{\textsc{BBC}}  & \multicolumn{2}{c}{\textsc{News20}}& \multicolumn{2}{c}{\textsc{BBC}}\\
 		& & Accuracy & F1-Score &Accuracy & F1-Score& Accuracy & F1-Score\\
 		\hline\\[-1em]
 		\textsc{Bi-LSTM (Target Only)} & & 65.17 & 0.6328& 91.33& 0.9122& 84.22& 0.8395\\
 		\textsc{Instance-Infused Bi-LSTM} & & 76.44 & 0.7586 &95.35 & 0.9531 & 88.78 & 0.8855\\ 
 		\textsc{Instance-Infused Bi-LSTM} (with penalty function) & & \textbf{78.29} & \textbf{0.7773}& \textbf{96.09}&  \textbf{0.9619}& \textbf{91.56}& \textbf{0.9100}\\ 
 		\hline
 	\end{tabular}\\
 	\vspace*{1mm}
 	\caption{Classification accuracies and F1-Scores for news arcticle classifications for different source and target domains. The first row corresponds to the baseline performance trained on only the target dataset. The following two rows shows the performance of instance-infusion method with and without the usage of penalty function. In all the three cases, our approach outperforms the baselines by a significant margin.}
 	\label{tab:result}
 \end{table*}

\subsection{Setup}
All experiments were carried on a Dell Precision Tower 7910 server with Quadro M5000 GPU with 8 GB of memory. The models were trained using the Adam's Optimizer~\cite{kingma2014adam} in a stochastic gradient descent~\cite{bottou2010large} fashion. The models were implemented using PyTorch~\cite{tensorflow2015-whitepaper}. The word embeddings were randomly initialized and trained along with the model. For testing purposes the algorithm was tested using 10-fold cross-validation scheme. The learning rate is regulated over the training epochs, it is decreased to 0.3 times its previous value after every 10 epochs.  The relevant hyper-parameters are listed in Table~\ref{tab:hyperparameters}. 

\subsection{Results}
\label{result}
Table~\ref{tab:result} shows the detailed results of our approach for all the datasets. The source and target datasets are chosen in a manner such that the source dataset is able to provide relevant information. 20Newsgroups contains news articles from all categories, so a good choice for source dataset is BBC which also encompasses articles from similar broad categories. For the same reason BBC also has 20Newsgroups as its source dataset. BBC Sports focuses on sports articles, BBC is chosen as the source dataset as the news articles share a common domain (articles come from same news media BBC).

For the proper functioning of the model, the final layer of the instance-infused model is replaced while the rest of the network is inherited from the pre-trained \textit{target only model}. We have shown improvements over the baseline by a high margin for all datasets, shown in Table~\ref{tab:result}. For 20Newsgroups the improvement over baseline model is 12\%, BBC and BBC Sports datasets show an improvement of around 5\%. As mentioned earlier, our approach is independent of the sentence encoder being used. Instead of bi-LSTM  any other model can be used. 
As the proposed approach is independent of the source encoding procedure and the source instance embeddings are kept constant during the training procedure, we can incorporate source instances from multiple datasets simultaneously. In the subsequent experimental setups, we try varying setups to prove the robustness and efficacy of our proposed model.

\begin{table}[h!]
	\centering
	\begin{tabular}{ c c c c}
		\hline 
		&&& \\[-1em]
		\textsc{Dataset} & \textsc{Accuracy} & \textsc{F1-Score} &  \textsc{Source Dataset}\\
		\hline
		&&& \\[-1em]
		\textsc{News20} &  77.51 & 0.7707  & \textsc{News20} \\ 
		\textsc{BBC} & 96.17 & 0.9606 & \textsc{BBC}\\ 
		\textsc{BBC Sports} & 90.63 & 0.8931 & \textsc{BBC Sports}\\ 
		\hline
	\end{tabular}\\
	\vspace*{1mm}
	\caption{Test Accuracy for proposed model using instances from the same target dataset}
	\label{samesource}
\end{table}

\textbf{Instance Infusion from Same Dataset}.
In this section, we study the results of using the pre-trained target dataset as the source for instance retrieval. This setting is same as the conventional instance-based learning setup. However, our approach not only uses the instance based information, but also leverage the macro statistics of the target dataset. As our proposed model is independent of the source dataset training scheme, we use the pre-trained target instances for source retrieval. 
The intuition behind this experimental setup is that instances from the same dataset is also useful in modeling other instances especially when a class imbalance exists in the target dataset. In this experimental setup, the \textit{nearest neighbour retrieved is ignored} as it would be same as the instance sample being modeled during training. The performance on the three news classification datasets is shown in Table~\ref{samesource}.

\begin{figure*}[h!]
   	\centering
   	\subfloat[]{\includegraphics[scale=0.7]{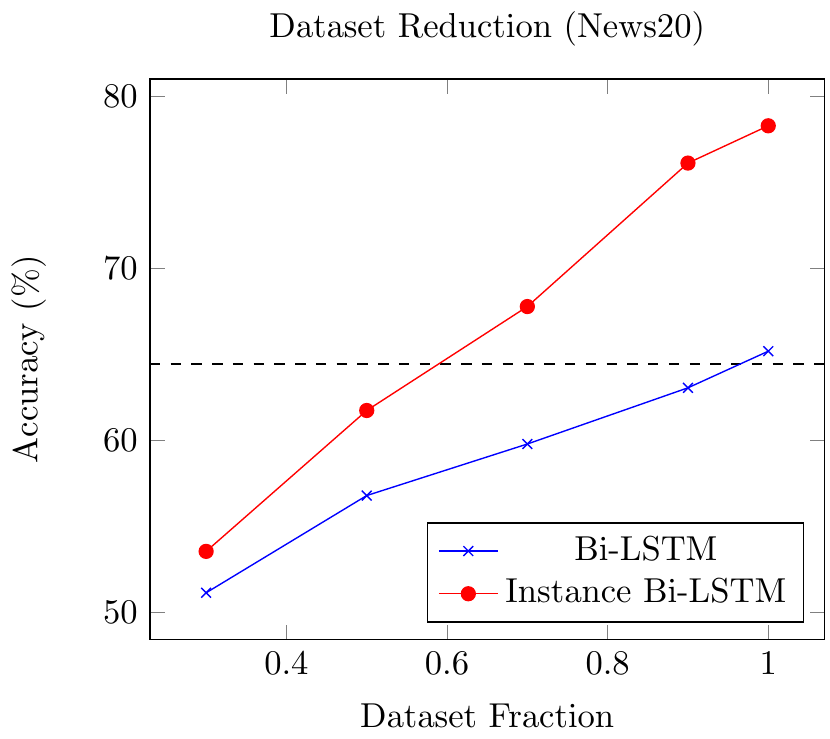}}\hspace*{0.3cm}
   	\subfloat[]{\includegraphics[scale=0.7]{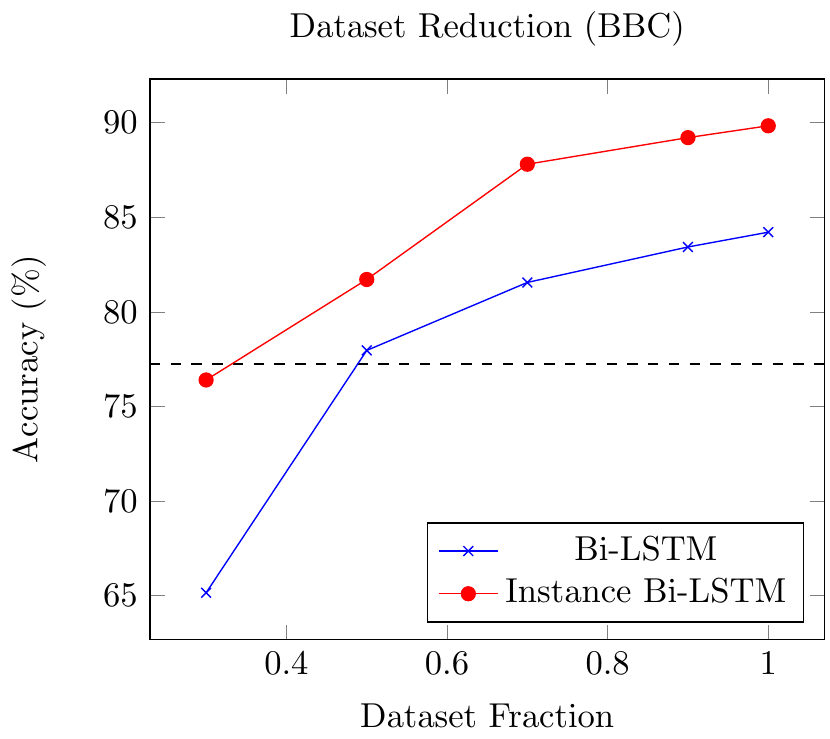}}\hspace*{0.3cm}
   	\subfloat[]{\includegraphics[scale=0.7]{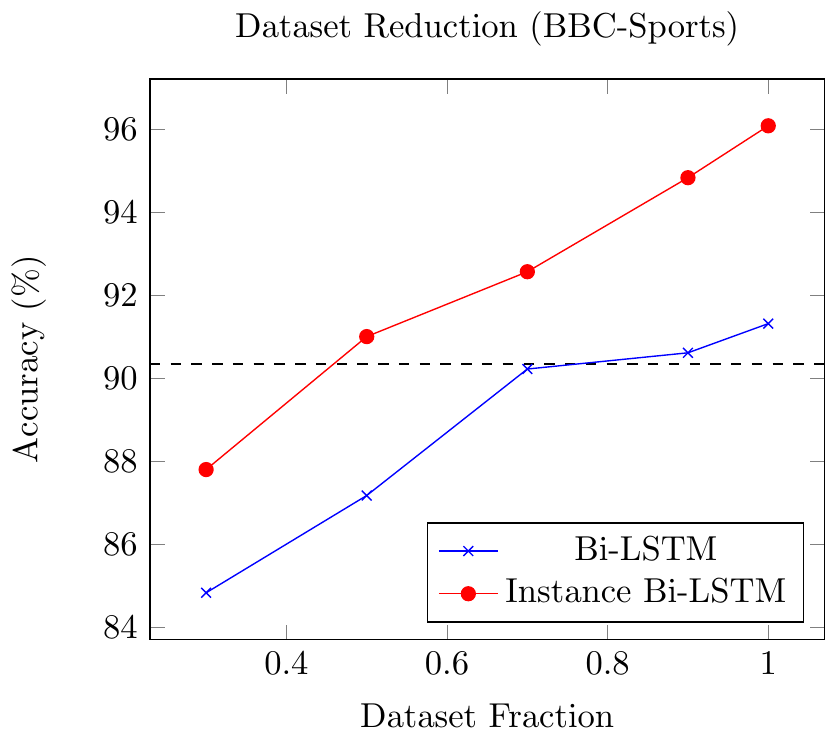}}
   	\caption{Accuracy Plot over dataset fractions for baseline and proposed model for (a) News20 (b) BBC (c) BBC Sports datasets}
   	\label{reduction}
   \end{figure*}

\textbf{Target Dataset Reduction with Single Source}.
 In this section, we discuss a set of experiments performed to support our hypothesis that the proposed model is capable for reduction of labeled instances in a dataset. In these set of experiments, we show that the cross-dataset augmented models perform significantly better than baseline models when varying fractions of the training data is used. Figure~\ref*{reduction} shows the variation of \textit{instance-infused bi-LSTM} and  \textit{bi-LSTM (target-only)} performance for 20Newsgroups, BBC and BBC Sports datasets. In these set of experiments 20Newsgroups had BBC, BBC had 20Newsgroup and BBC Sports had BBC as source dataset. As shown in the plot, 0.3, 0.5, 0.7, 0.9 and 1.0 fraction of the dataset are used for performance analysis. For all dataset fractions, the proposed model beats the baseline by a significant margin. The dashed line in the plots indicates the baseline model performance with 100\% target dataset support. It is observed that the performance of instance-infused bi-LSTM with 70\% dataset, is better than the baseline model trained on the entire dataset. This observation shows that our proposed approach is successful in reducing the dependency on the training examples by at least 30\% across all datasets in our experiments.

  \begin{table}[h!]
  	\centering
  	\begin{tabular}{ c c c c c}
  		\hline 
  		&&&& \\[-1em]
  		\multirow{2}{*}{\textsc{Dataset}} & \multicolumn{2}{c}{\textsc{Single Source}} & \multicolumn{2}{c}{\textsc{Multiple Sources}}\\
  		& \textsc{Accuracy} & \textsc{F1-Score} &  \textsc{Accuracy} & \textsc{F1-Score}\\
  		\hline
  		&&&& \\[-1em]
  		\textsc{News20} & 61.72 & 0.6133 & 67.32 & 0.6650\\ 
  		\textsc{BBC} & 91.01& 0.9108& 91.41 & 0.9120 \\ 
  		\textsc{BBC Sports}  & 81.72 & 0.7990& 82.81 & 0.8027\\ 
  		\hline
  	\end{tabular}\\
  	\vspace*{1mm}
  	\caption{Test Accuracy for proposed model using instances from multiple source datasets with 50\% target dataset}
  	\label{multiplesource1}
  \end{table}
  
 \textbf{Target Dataset Reduction with Multiple Source}. In this section, we design an experimental setup in which only 0.5 fraction of the target dataset is utilized and study the influence of multiple source dataset infusion. Table~\ref{multiplesource1} compares the results, when single source and multiple source datasets are used for 50\% dataset fraction. The results improve as and when more source datasets are used in the infusion process. This can be effectively leveraged for improving the performance of very lean datasets, by heavily deploying large datasets as source. For the single source setup, the same source datasets are used as mentioned in Section~\ref{result}. In multiple source experiment setup, for a given target dataset the other two datasets are used as source.

   \begin{figure}[h!]
   	\centering
   	\subfloat[]{\includegraphics[scale=0.2]{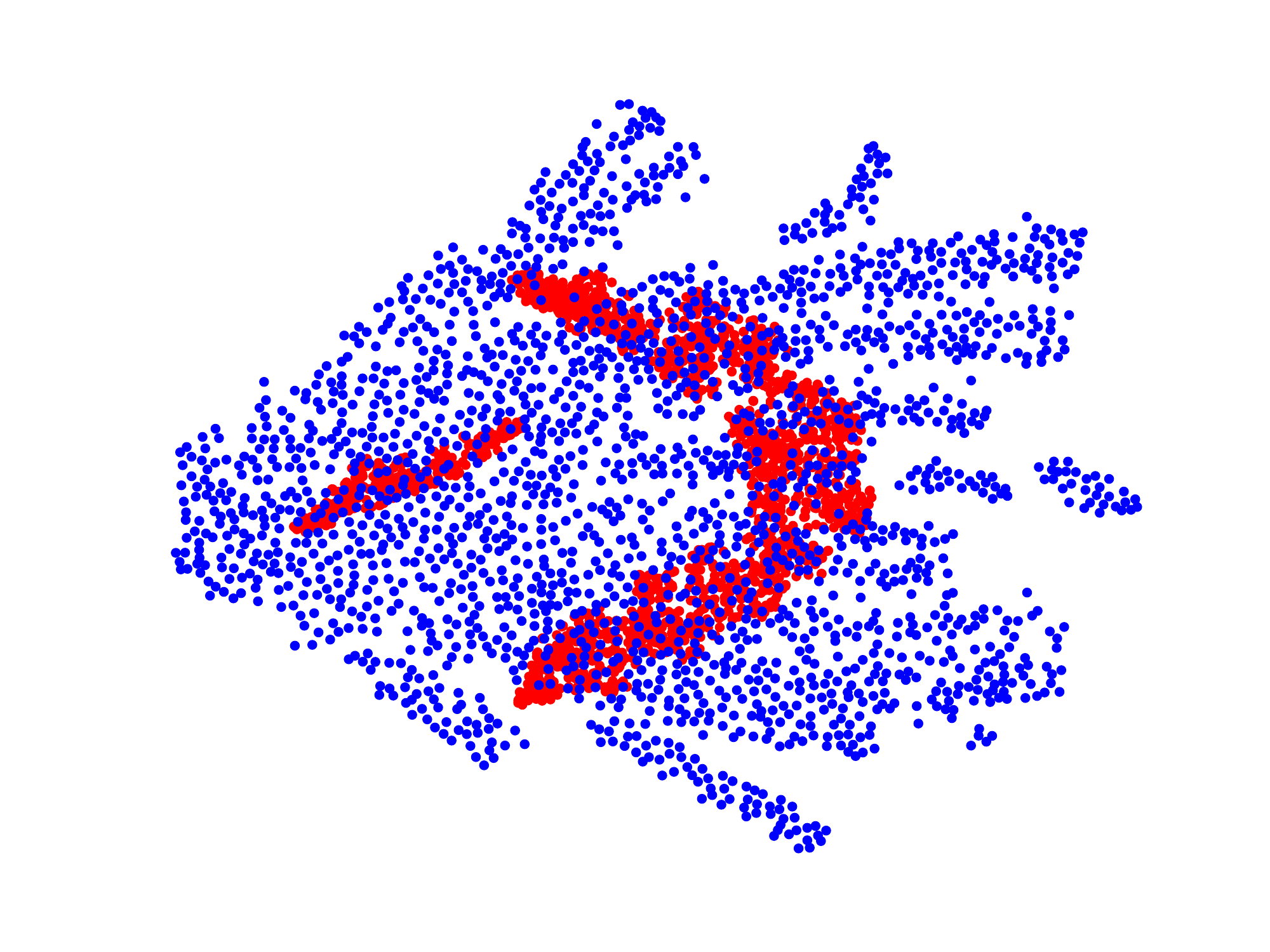}}\hspace*{0.3cm}
   	\subfloat[]{\includegraphics[scale=0.2]{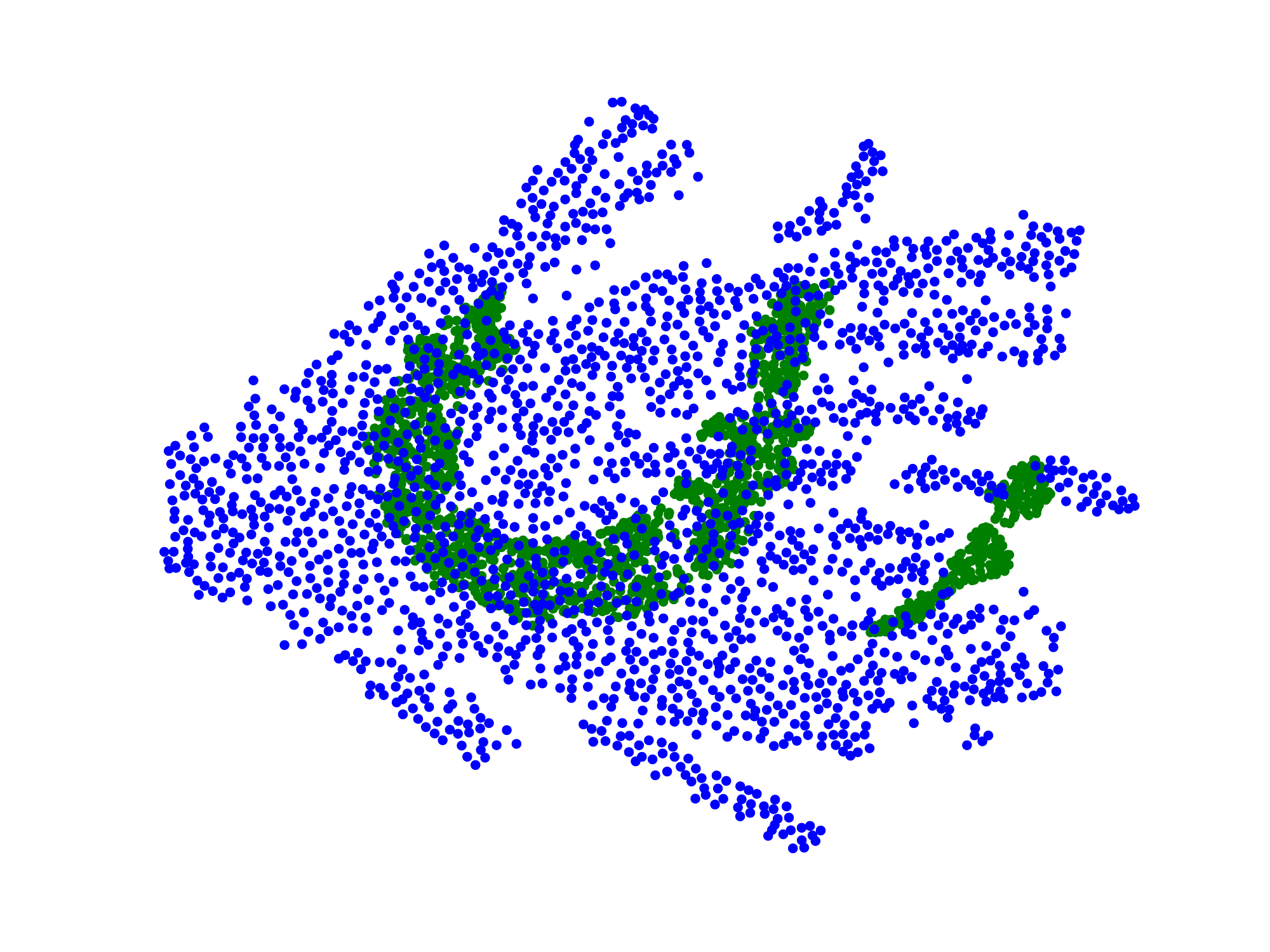}}\\%\hspace*{0.3cm}
   	\subfloat[]{\includegraphics[scale=0.2]{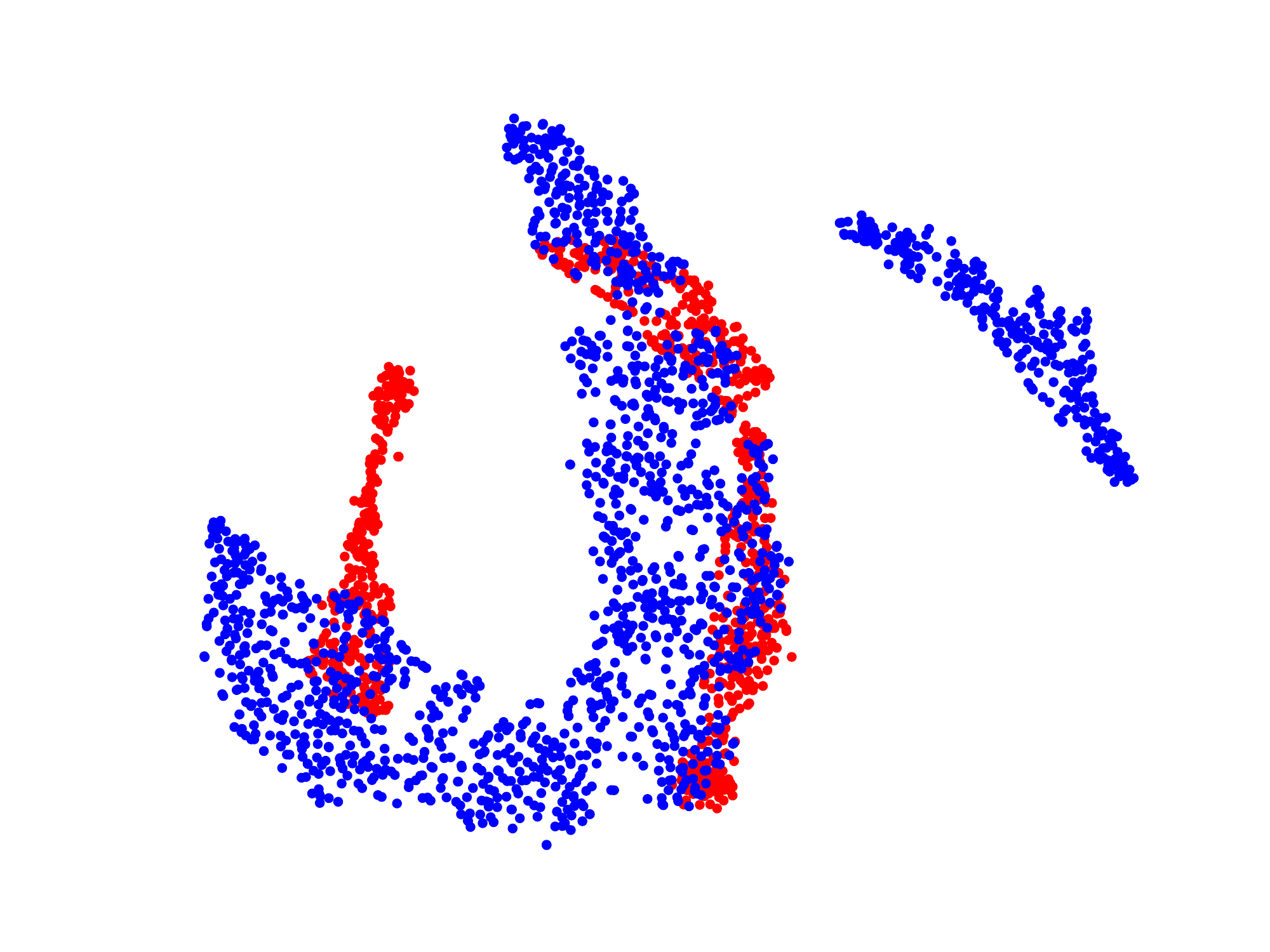}}\hspace*{0.3cm}
   	\subfloat[]{\includegraphics[scale=0.2]{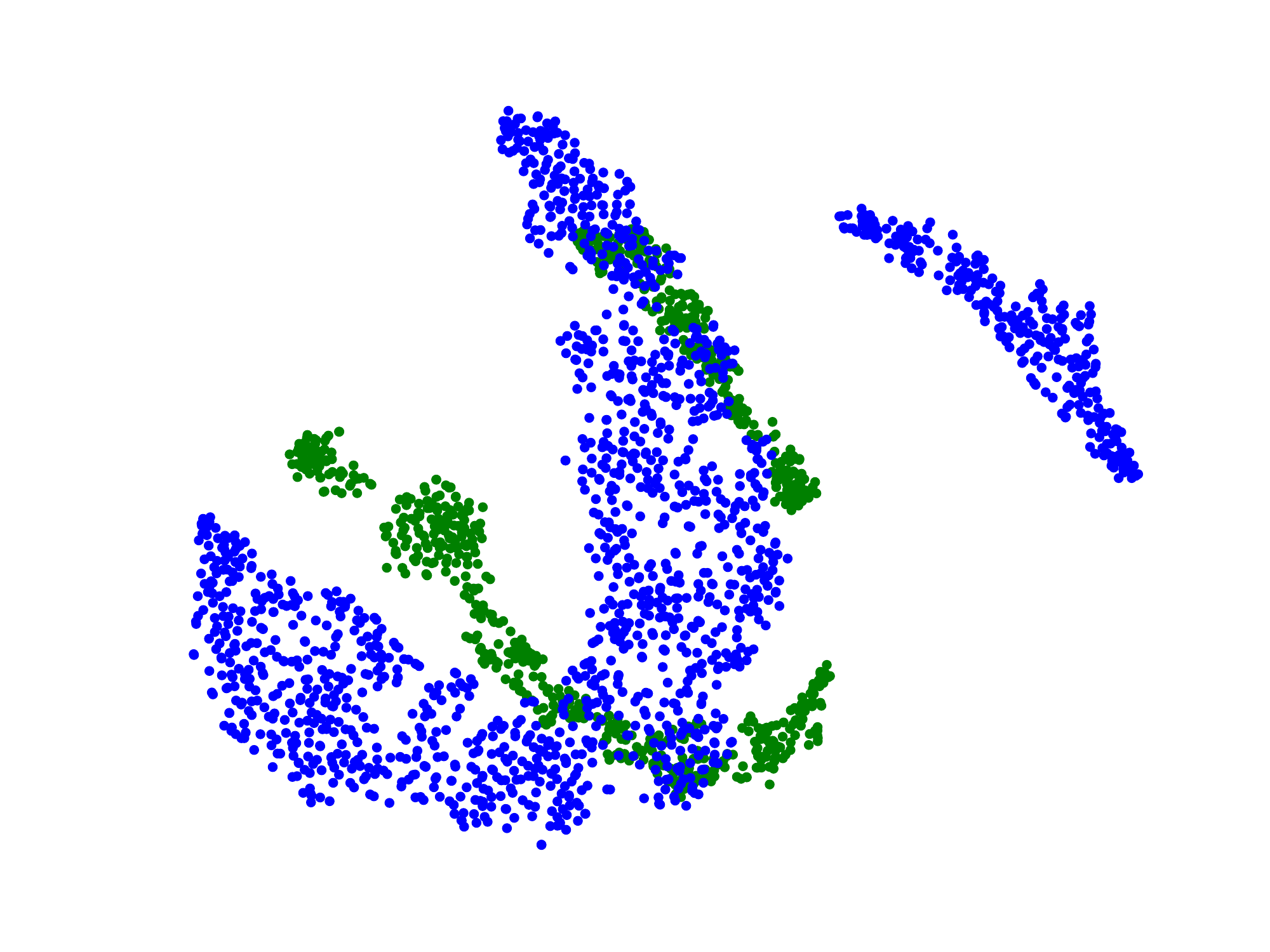}}
   	\caption{(a) and (b) show t-SNE visualizations of BBC as target dataset and News20 as source dataset.   (a) Plain embeddings of target BBC(in red) (b) Instance Infused embeddings of target BBC(in green). (c) and (d) show t-SNE visualizations of BBC Sports as target dataset and BBC as source dataset. (c) Plain embeddings of target BBC Sports (in red) (d) Instance Infused embeddings of target BBC Sports (in green).   }
   	\label{visualization}
   \end{figure}
 
\begin{figure}[h!]
     	\centering
     	\subfloat[]{\includegraphics[scale=0.2]{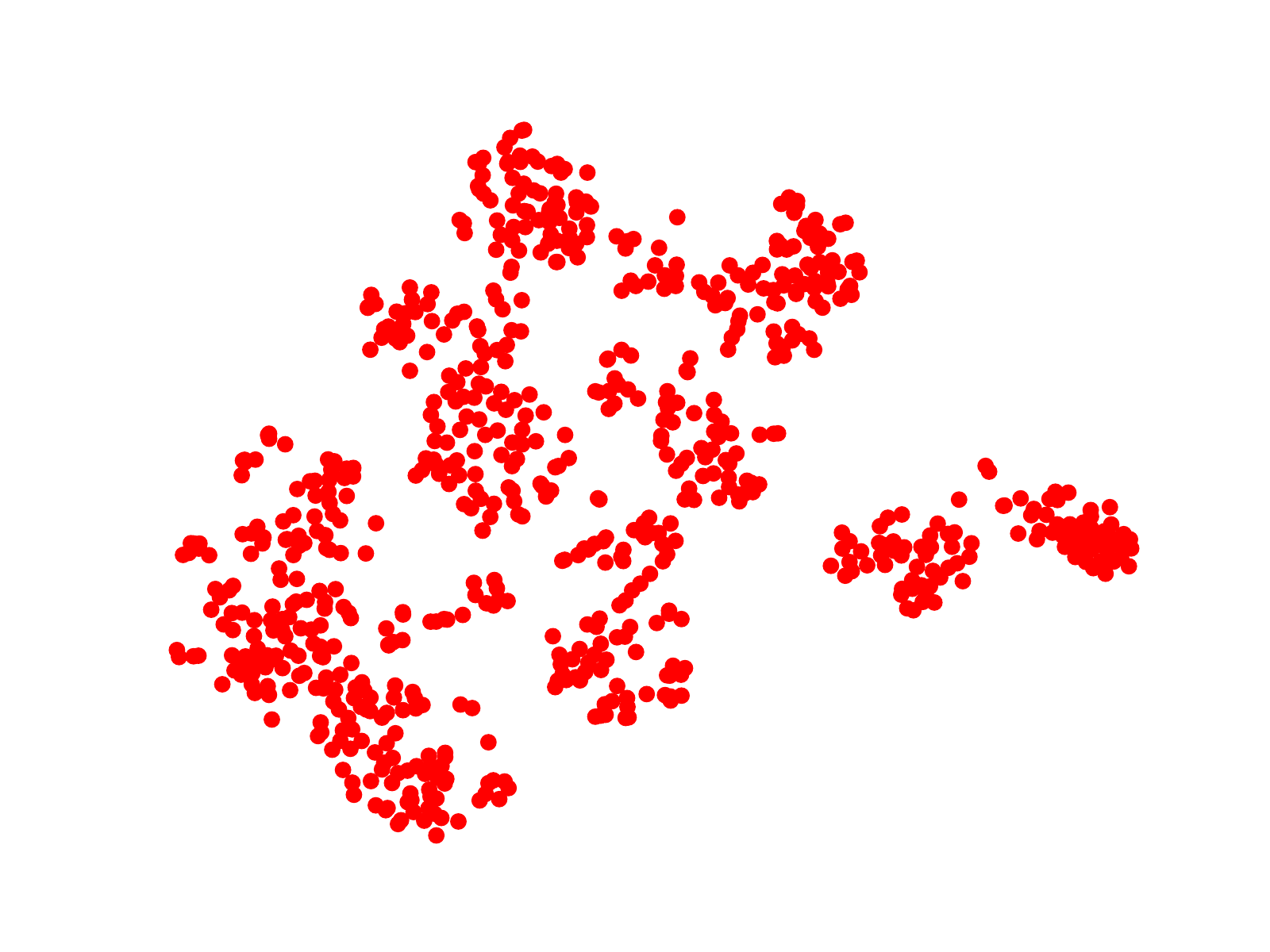}}\hspace*{0.3cm}
     	\subfloat[]{\includegraphics[scale=0.2]{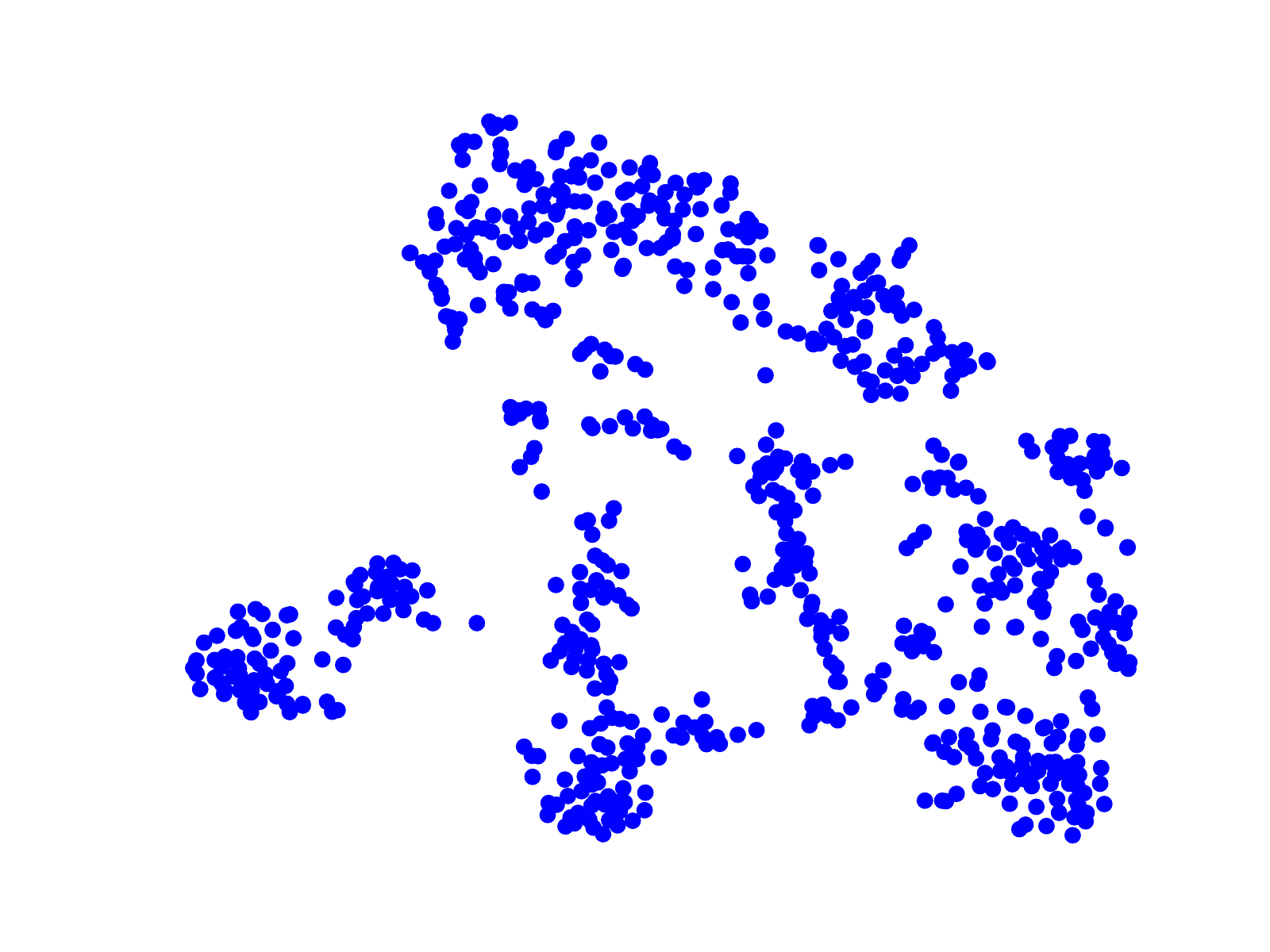}}\\
     	
     	\subfloat[]{\includegraphics[scale=0.2]{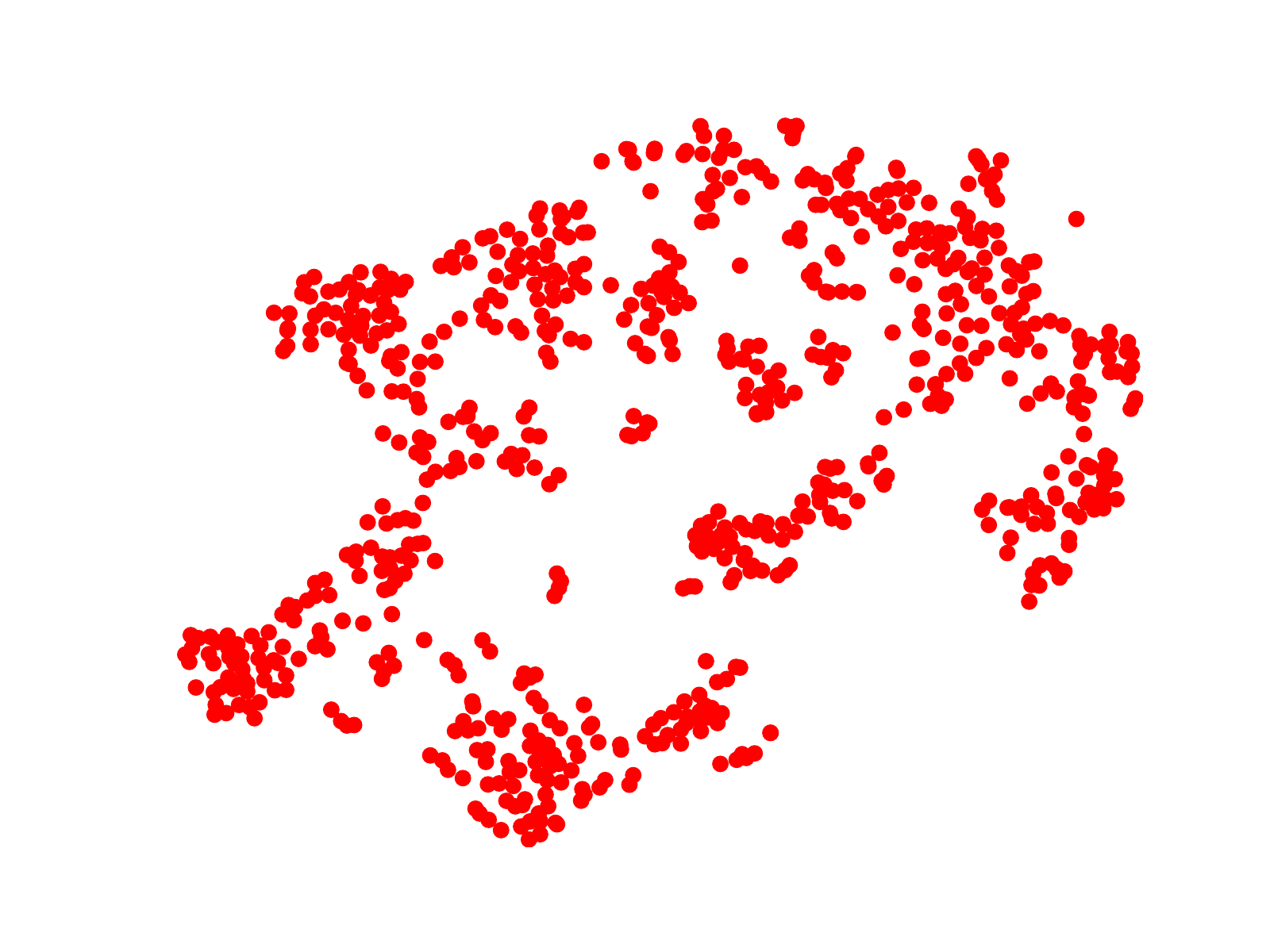}}\hspace*{0.3cm}
     	\subfloat[]{\includegraphics[scale=0.2]{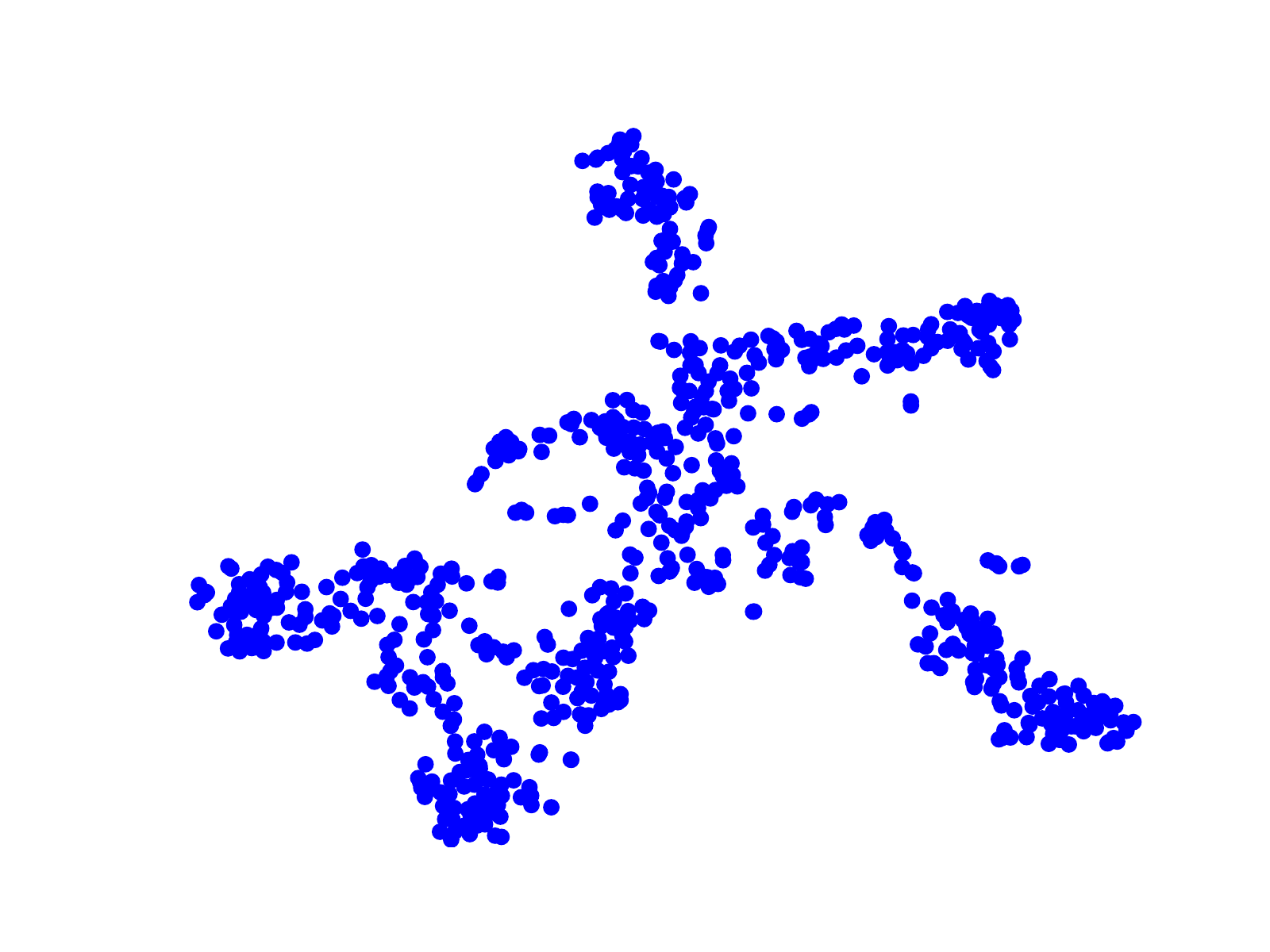}}\\

     	\subfloat[]{\includegraphics[scale=0.2]{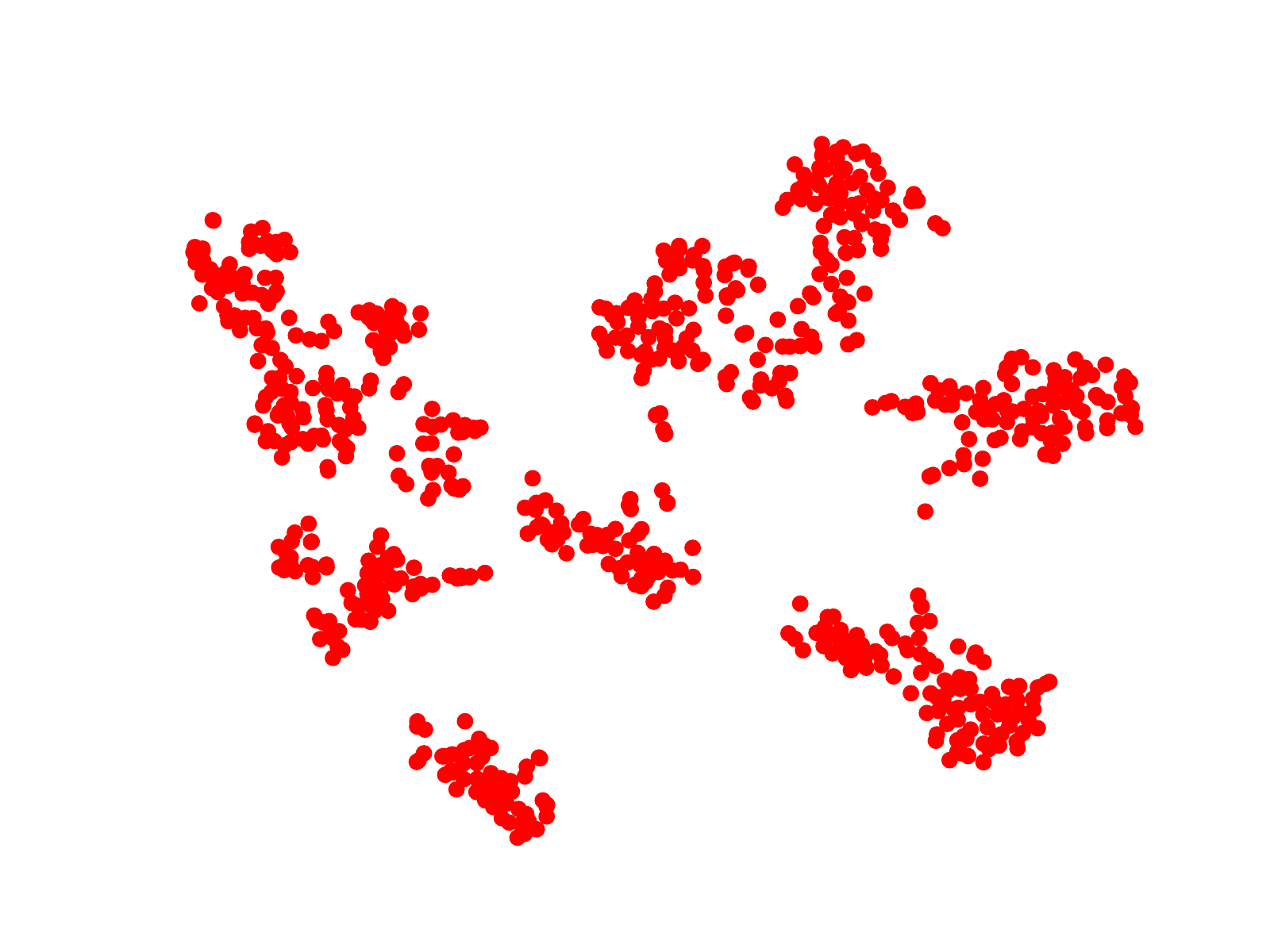}}\hspace*{0.3cm}
     	\subfloat[]{\includegraphics[scale=0.2]{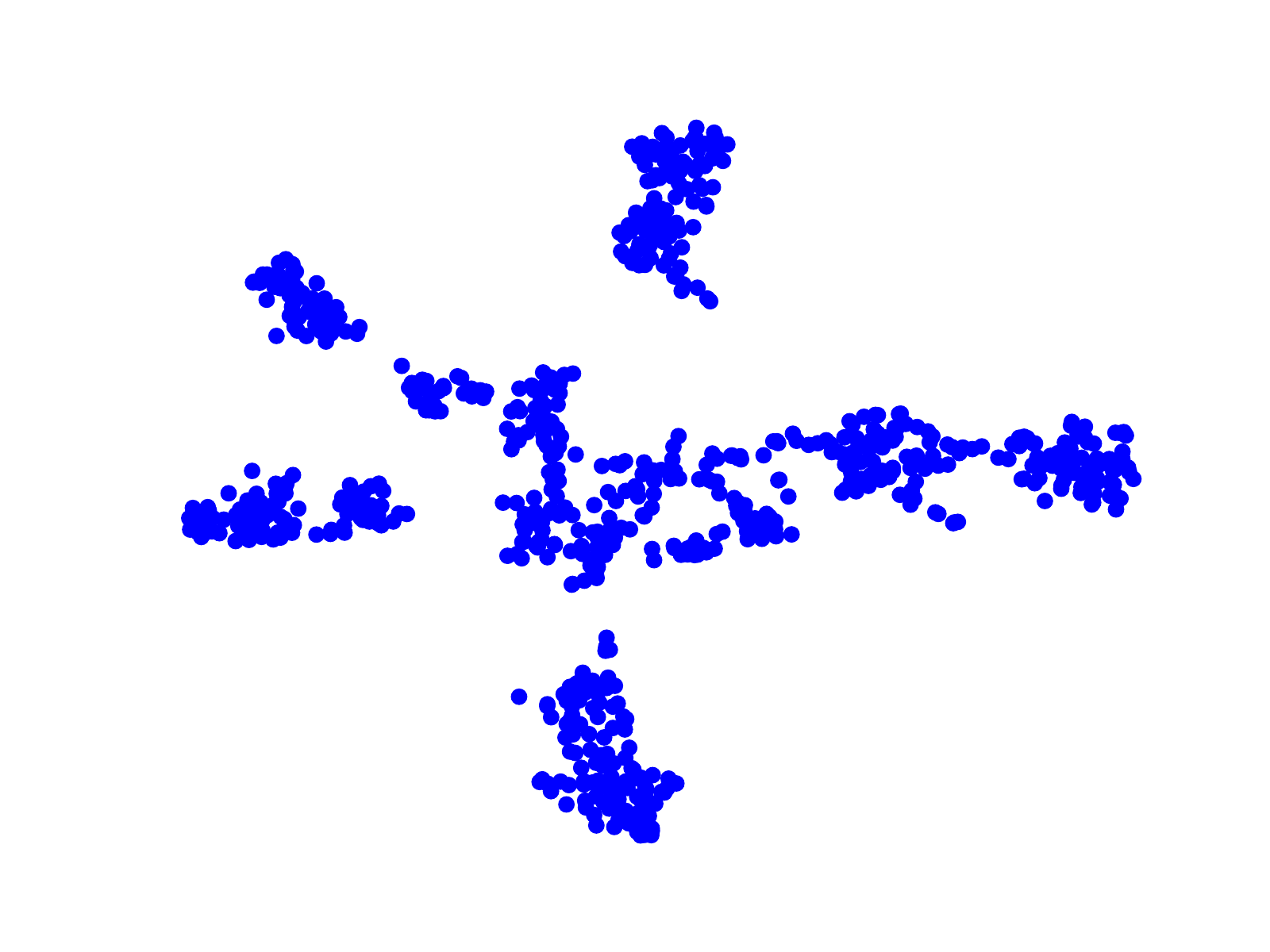}}\\
     	\caption{t-SNE visualization of latent space vectors of BBC Sports with News20 as source dataset for varying dataset fractions. Plain embeddings are denoted in red and Instance-infused embeddings are represented in blue. (a) \& (b) show embeddings for 30\% data fraction, (c) \& (d) for 50\% data, and (e) \& (f) for 70\% data.}
     	\label{visualization3}
     \end{figure}

 \textbf{Visualization}
 In this section we show some visualization of latent space embeddings obtained using \textit{bi-LSTM (target only)} and with \textit{instance
 	infusion}. Figure~\ref{visualization} shows the  
 t-SNE~\cite{maaten2008visualizing} visualization for target datasets
 BBC with source dataset 20Newsgroups and BBC Sports with source
 dataset BBC. Figure~\ref{visualization} (a) and (b), correspond to
 visualizations with BBC as target dataset and
 Figure~\ref{visualization} (c) and (d) correspond to visualizations
 with BBC Sports as target dataset. For Figure~\ref{visualization} (a)
 and (b), the source dataset embeddings of News20 are sparsified using
 \textit{instance clustering} (described in Section~\ref{clustering}) (number of clusters =
 2000) for better visualization. In the figure, the embeddings marked
 by blue denote source vector space, those represented by red denote
 \textit{bi-LSTM (target only)} embeddings and embeddings represented
 by green correspond to those from the \textit{instance-infused}
 model. Figure~\ref{visualization} shows that the embedding
 visualization change drastically using our model which in turn
 improves performance. It is visible from the figure that the latent
 vectors try to shape themselves in a manner so that the difference
 between the source and target distribution is reduced.  We show the instance infusion is in fact accelerating the learning procedure, by analyzing how the latent vector space representation change with varying training data fractions of the target dataset. 
 In Figure~\ref{visualization3}, the latent vector embeddings of BBC Sports dataset with News20 support is shown for 0.3 in (a) \& (b), 0.5 in (c) \& (d) and 0.7 in (e) \& (f), fraction of the target training dataset (BBC Sports).
 Figure~\ref{visualization3} (f) is the embeddings representation with 70\% data for which best performance (among the 6 visualizations) is observed.

 It is evident from the figure that even with 30\% and 50\% of the data \textit{instance infusion} tries to make the embedding distribution similar to Figure~\ref{visualization3} (f) as seen in Figure~\ref{visualization3} (b) and (d), when the \textit{bi-LSTM (target-only)} instances representations in Figure~\ref{visualization3} (a) and (c) are quite different. This illustrates that by instance infusion the latent space evolves faster to the better performing shape compared to the scenario where no instance infusion is done.

   \begin{table*}[h!]
   	\centering
   	\begin{tabular}{ p{4cm} c c c c c c}
   		\hline \\[-0.5em]
   		\multirow{2}{*}{\textsc{Model}} &\multicolumn{2}{c}{\textsc{News20}}  & \multicolumn{2}{c}{\textsc{BBC}}& \multicolumn{2}{c}{\textsc{BBC Sports}}\\
   		& Accuracy & F1-Score &Accuracy & F1-Score& Accuracy & F1-Score\\ \\[-.5em]
   		\hline\\[-1em]
   		kNN-ngrams & 35.25 &  0.3566 & 74.61 & 0.7376 & 94.59 & 0.9487\\
   		Multinomial NB-bigram & \textbf{79.21} & \textbf{0.7841} & 95.96 & 0.9575 & 95.95 & 0.9560\\
   		SVM-bigram & 75.04 & 0.7474 & 94.83 & 0.9456 & 93.92 & 0.9393\\
   		SVM-ngrams & 78.60 & 0.7789 & 95.06 & 0.9484 & 95.95 & 0.9594\\
   		Random Forests-bigram & 69.01 & 0.6906 & 87.19 &  0.8652 & 85.81 & 0.8604\\
   		Random Forests-ngrams & 78.36 & 0.7697 & 94.83 & 0.9478 & 94.59 & 0.9487\\
   		Random Forests- tf-idf & 78.6 & 0.7709 & 95.51 & 0.9547 & \textbf{96.62} & \textbf{0.9660}\\ \\[-1em]
   		\hline\\[-1em]
   		Bi-LSTM & 65.17 & 0.6328& 91.33& 0.9122& 84.22& 0.8395\\
   		Instance-Infused Bi-LSTM & 78.29 & 0.7773& \textbf{96.09}&  \textbf{0.9619}& 91.56& 0.9100\\ 
   		\hline
   	\end{tabular}
   	\vspace*{1mm}
   	\caption{Comparison of results using other learning schemes on News20, BBC and BBC Sports datasets. The proposed model using a deep learning model as a baseline achieves competitive performance for all the three datasets.}
   	\label{tab:comparison}
   \end{table*}

  \textbf{Comparative Study}.
  Table~\ref{tab:comparison} gives the experimental results for our proposed approach, baselines and other conventional learning techniques on the 20 Newsgroups, BBC and BBC Sports datasets. Literature involving these datasets mainly focus on non-deep learning based approaches. Thereby, we compare our results with some popular conventional learning techniques. The experiments involving conventional learning were performed using \textit{scikit-learn}~\cite{scikit-learn} library in Python\footnote{https://www.python.org/}. For the k-NN-ngram experiments, the number of nearest neighbours $k$ was set to 5. In Table~\ref{tab:comparison}, the models studied are Multinomial Naive Bayes~\cite{kibriya2004multinomial}, $k$-nearest neighbour classifier~\cite{cunningham2007k}, Support Vector Machine~\cite{suykens1999least} (SVM) and Random Forests Classifier~\cite{breiman2001random}. The input vectors were initialized using n-grams~\cite{brown1992class}, bi-gram or term frequency-inverse document frequency (tf-idf). For the mentioned datasets, conventional models outperform our baseline \textit{Bi-LSTM} model, however upon \textit{instance infusion} the deep learning based model is able to achieve competitive performance across all datasets. Moreover by  instance infusion the simple bi-LSTM model approaches the classical models in performance on News20 and BBC Sports dataset, whereas on BBC Dataset the proposed instance infused bi-LSTM model beats all the mentioned models.

\subsection{Related Work} 
 The goal of this work is to efficiently utilize knowledge extant in a secondary dataset of a similar domain that can be closely linked
 to transfer learning and domain adaptation. Domain adaptation and
 transfer learning~\cite{pan2010survey} aim at utilizing domain
 knowledge of one task for another task and also learning task
 independent representations. This also reduces the dependency of
 learning algorithms on labeled data. The challenge in these tasks lie
 in learning a representation that can reduce the discrepancy in
 probability distributions across domains. 
 
 There has been an array of work in the field of domain adaptation, we
 mention a few relevant works here. One of the popular work in this
 field is Domain Adaptive Networks (DAN)~\cite{ghifary2014domain},
 which penalizes the learning algorithm using Maximum Mean Discrepancy
 (MMD)  metric used to compute the distance between source and target
 distribution.~\cite{glorot2011domain} uses a two-step approach using a
 stacked autoencoder architecture to reduce the discrepancy between the
 source and target domain.~\cite{long2016unsupervised} uses residual
 networks for unsupervised domain knowledge transfer. With the advent
 of deep networks, it is now easier to learn latent representations
 which is accessible across various domains.~\cite{donahue2014decaf}
 studies this feature of deep networks of learning latent
 visualizations and how they vary across domain specific
 tasks.~\cite{wanggleaning} uses an active learning method for querying
 most informative instances from multiple
 domain.~\cite{collobert2011natural} uses deep learning approaches for
 multi-task learning for a variety of tasks in natural language
 processing domain. 
 
 The setup of our framework is different from the conventional domain
 adaptation based methods. We are given a source domain
 $\mathcal{D}_s = \{(x^s_i, y^s_i)\}^{n_s}_{i=1}$ of $n_s$ and a target
 domain $\mathcal{D}_t = \{x^t_j, y^t_i\}^{n_t}_{j=1}$ of $n_t$ labeled
 instances. We aim to utilize the pre-trained model on
 $\mathcal{D}_s$ to access its latent vectors $\mathcal{Z}_s =
 \{z^s_i\}^{n_s}_{i=1}$. In our method, instances from the target
 dataset queries similar latent vector instances from the source
 dataset thereby formulating an instance retrieval based transfer
 learning policy.
 
\section{Conclusion \& Future Work}
\label{future}
In this work we posit that the infusion of instance level local information along with macro statistics of the dataset, can significantly improve the performance of learning algorithms. Through extensive experimentation, we have shown that our approach can improve the performance of learning models significantly. The improvement in performance shows that this approach has potential and very useful in cases where the dataset is very lean.  Although instance based learning is very well studied in machine learning literature, this has rarely been used in a deep learning setup for knowledge transfer. Moreover, approaches to exploit instance level local information and macro statistics of the dataset is an exciting topic to embark on. One thread of work which can be pursued to improve our setup, is to enhance the search paradigm to retrieve instances to reduce latency during training. In this work, we have shown extensive experiments where our method reduces dependency on labeled data, however this work may be extended to analyze performance in a purely unsupervised setup. Improved feature modification techniques can be augmented along with the search module in order to enhance the query formulation. In this work, we also assumed that the datasets share a common domain, in future work means to tackle domain discrepancy needs to be formulated to incorporate instances from a range of datasets.

% Although we have formulated the proposed model using bi-LSTM, as mentioned before, the approach is generic and can be applied with any other architecture as base. Detailed experimentation for various problems in NLP like Natural Language Inference~\cite{maccartney2009natural}, with the state of the art achitectures is one thread of work that should be pursued. Locality Sensitive Hashing is trained offline in the architecture after the pre-training stage to create the embeddings. A neural network based hashing scheme~\cite{mu2017deep} can be devised, which helps to train the model end to end, rather than the step wise procedure we have used. This will also help to learn hashing function much amenable for the final classification task. Experimental results show that the approach has a lot of potential, and there are many work directions that can be embarked on.

\bibliographystyle{ACM-Reference-Format}
\bibliography{CDL-arxiv}


\begin{thebibliography}{33}

%%% the Web address.

\ifx \showCODEN    \undefined \def \showCODEN     #1{\unskip}     \fi
\ifx \showDOI      \undefined \def \showDOI       #1{#1}\fi
\ifx \showISBNx    \undefined \def \showISBNx     #1{\unskip}     \fi
\ifx \showISBNxiii \undefined \def \showISBNxiii  #1{\unskip}     \fi
\ifx \showISSN     \undefined \def \showISSN      #1{\unskip}     \fi
\ifx \showLCCN     \undefined \def \showLCCN      #1{\unskip}     \fi
\ifx \shownote     \undefined \def \shownote      #1{#1}          \fi
\ifx \showarticletitle \undefined \def \showarticletitle #1{#1}   \fi
\ifx \showURL      \undefined \def \showURL       {\relax}        \fi
% invisible to TeX
\providecommand\bibfield[2]{#2}
\providecommand\bibinfo[2]{#2}
\providecommand\natexlab[1]{#1}
\providecommand\showeprint[2][]{arXiv:#2}

\bibitem[\protect\citeauthoryear{Abadi, Agarwal, Barham, Brevdo, Chen, and
  et~al.}{Abadi et~al\mbox{.}}{2015}]%
        {tensorflow2015-whitepaper}
\bibfield{author}{\bibinfo{person}{Mart\'{\i}n Abadi}, \bibinfo{person}{Ashish
  Agarwal}, \bibinfo{person}{Paul Barham}, \bibinfo{person}{Eugene Brevdo},
  \bibinfo{person}{Zhifeng Chen}, {and} \bibinfo{person}{Craig~Citro et al.}}
  \bibinfo{year}{2015}\natexlab{}.
\newblock \bibinfo{title}{{TensorFlow}: Large-Scale Machine Learning on
  Heterogeneous Systems}.
\newblock   (\bibinfo{year}{2015}).
\newblock
\urldef\tempurl%
\url{http://tensorflow.org/}
\showURL{%
\tempurl}
\newblock
\shownote{Software available from tensorflow.org.}


\bibitem[\protect\citeauthoryear{Aggarwal}{Aggarwal}{2014}]%
        {aggarwal2014instance}
\bibfield{author}{\bibinfo{person}{Charu~C Aggarwal}.}
  \bibinfo{year}{2014}\natexlab{}.
\newblock \bibinfo{title}{Instance-Based Learning: A Survey.}
\newblock   (\bibinfo{year}{2014}).
\newblock


\bibitem[\protect\citeauthoryear{Bawa, Condie, and Ganesan}{Bawa
  et~al\mbox{.}}{2005}]%
        {bawa2005lsh}
\bibfield{author}{\bibinfo{person}{Mayank Bawa}, \bibinfo{person}{Tyson
  Condie}, {and} \bibinfo{person}{Prasanna Ganesan}.}
  \bibinfo{year}{2005}\natexlab{}.
\newblock \showarticletitle{LSH forest: self-tuning indexes for similarity
  search}. In \bibinfo{booktitle}{\emph{Proceedings of the 14th international
  conference on World Wide Web}}. ACM, \bibinfo{pages}{651--660}.
\newblock


\bibitem[\protect\citeauthoryear{Bishop}{Bishop}{2006}]%
        {bishop2006pattern}
\bibfield{author}{\bibinfo{person}{Christopher~M Bishop}.}
  \bibinfo{year}{2006}\natexlab{}.
\newblock \bibinfo{booktitle}{\emph{Pattern recognition and machine learning}}.
\newblock \bibinfo{publisher}{springer}.
\newblock


\bibitem[\protect\citeauthoryear{Bottou}{Bottou}{2010}]%
        {bottou2010large}
\bibfield{author}{\bibinfo{person}{L{\'e}on Bottou}.}
  \bibinfo{year}{2010}\natexlab{}.
\newblock \showarticletitle{Large-scale machine learning with stochastic
  gradient descent}.
\newblock In \bibinfo{booktitle}{\emph{Proceedings of COMPSTAT'2010}}.
  \bibinfo{publisher}{Springer}, \bibinfo{pages}{177--186}.
\newblock


\bibitem[\protect\citeauthoryear{Breiman}{Breiman}{2001}]%
        {breiman2001random}
\bibfield{author}{\bibinfo{person}{Leo Breiman}.}
  \bibinfo{year}{2001}\natexlab{}.
\newblock \showarticletitle{Random forests}.
\newblock \bibinfo{journal}{\emph{Machine learning}} \bibinfo{volume}{45},
  \bibinfo{number}{1} (\bibinfo{year}{2001}), \bibinfo{pages}{5--32}.
\newblock


\bibitem[\protect\citeauthoryear{Brown, Desouza, Mercer, Pietra, and Lai}{Brown
  et~al\mbox{.}}{1992}]%
        {brown1992class}
\bibfield{author}{\bibinfo{person}{Peter~F Brown}, \bibinfo{person}{Peter~V
  Desouza}, \bibinfo{person}{Robert~L Mercer}, \bibinfo{person}{Vincent J~Della
  Pietra}, {and} \bibinfo{person}{Jenifer~C Lai}.}
  \bibinfo{year}{1992}\natexlab{}.
\newblock \showarticletitle{Class-based n-gram models of natural language}.
\newblock \bibinfo{journal}{\emph{Computational linguistics}}
  \bibinfo{volume}{18}, \bibinfo{number}{4} (\bibinfo{year}{1992}),
  \bibinfo{pages}{467--479}.
\newblock


\bibitem[\protect\citeauthoryear{Collobert, Weston, Bottou, Karlen,
  Kavukcuoglu, and Kuksa}{Collobert et~al\mbox{.}}{2011}]%
        {collobert2011natural}
\bibfield{author}{\bibinfo{person}{Ronan Collobert}, \bibinfo{person}{Jason
  Weston}, \bibinfo{person}{L{\'e}on Bottou}, \bibinfo{person}{Michael Karlen},
  \bibinfo{person}{Koray Kavukcuoglu}, {and} \bibinfo{person}{Pavel Kuksa}.}
  \bibinfo{year}{2011}\natexlab{}.
\newblock \showarticletitle{Natural language processing (almost) from scratch}.
\newblock \bibinfo{journal}{\emph{Journal of Machine Learning Research}}
  \bibinfo{volume}{12}, \bibinfo{number}{Aug} (\bibinfo{year}{2011}),
  \bibinfo{pages}{2493--2537}.
\newblock


\bibitem[\protect\citeauthoryear{Cunningham and Delany}{Cunningham and
  Delany}{2007}]%
        {cunningham2007k}
\bibfield{author}{\bibinfo{person}{Padraig Cunningham} {and}
  \bibinfo{person}{Sarah~Jane Delany}.} \bibinfo{year}{2007}\natexlab{}.
\newblock \showarticletitle{k-Nearest neighbour classifiers}.
\newblock \bibinfo{journal}{\emph{Multiple Classifier Systems}}
  \bibinfo{volume}{34} (\bibinfo{year}{2007}), \bibinfo{pages}{1--17}.
\newblock


\bibitem[\protect\citeauthoryear{Dai, Yang, Xue, and Yu}{Dai
  et~al\mbox{.}}{2008}]%
        {dai2008self}
\bibfield{author}{\bibinfo{person}{Wenyuan Dai}, \bibinfo{person}{Qiang Yang},
  \bibinfo{person}{Gui-Rong Xue}, {and} \bibinfo{person}{Yong Yu}.}
  \bibinfo{year}{2008}\natexlab{}.
\newblock \showarticletitle{Self-taught clustering}. In
  \bibinfo{booktitle}{\emph{Proceedings of the 25th international conference on
  Machine learning}}. ACM, \bibinfo{pages}{200--207}.
\newblock


\bibitem[\protect\citeauthoryear{Daume~III and Marcu}{Daume~III and
  Marcu}{2006}]%
        {daume2006domain}
\bibfield{author}{\bibinfo{person}{Hal Daume~III} {and} \bibinfo{person}{Daniel
  Marcu}.} \bibinfo{year}{2006}\natexlab{}.
\newblock \showarticletitle{Domain adaptation for statistical classifiers}.
\newblock \bibinfo{journal}{\emph{Journal of Artificial Intelligence Research}}
   \bibinfo{volume}{26} (\bibinfo{year}{2006}), \bibinfo{pages}{101--126}.
\newblock


\bibitem[\protect\citeauthoryear{Donahue, Jia, Vinyals, Hoffman, Zhang, Tzeng,
  and Darrell}{Donahue et~al\mbox{.}}{2014}]%
        {donahue2014decaf}
\bibfield{author}{\bibinfo{person}{Jeff Donahue}, \bibinfo{person}{Yangqing
  Jia}, \bibinfo{person}{Oriol Vinyals}, \bibinfo{person}{Judy Hoffman},
  \bibinfo{person}{Ning Zhang}, \bibinfo{person}{Eric Tzeng}, {and}
  \bibinfo{person}{Trevor Darrell}.} \bibinfo{year}{2014}\natexlab{}.
\newblock \showarticletitle{Decaf: A deep convolutional activation feature for
  generic visual recognition}. In \bibinfo{booktitle}{\emph{International
  conference on machine learning}}. \bibinfo{pages}{647--655}.
\newblock


\bibitem[\protect\citeauthoryear{Gao, Jagadish, Lu, and Ooi}{Gao
  et~al\mbox{.}}{2014}]%
        {gao2014dsh}
\bibfield{author}{\bibinfo{person}{Jinyang Gao},
  \bibinfo{person}{Hosagrahar~Visvesvaraya Jagadish}, \bibinfo{person}{Wei Lu},
  {and} \bibinfo{person}{Beng~Chin Ooi}.} \bibinfo{year}{2014}\natexlab{}.
\newblock \showarticletitle{DSH: data sensitive hashing for high-dimensional
  k-nnsearch}. In \bibinfo{booktitle}{\emph{Proceedings of the 2014 ACM SIGMOD
  international conference on Management of data}}. ACM,
  \bibinfo{pages}{1127--1138}.
\newblock


\bibitem[\protect\citeauthoryear{Ghifary, Kleijn, and Zhang}{Ghifary
  et~al\mbox{.}}{2014}]%
        {ghifary2014domain}
\bibfield{author}{\bibinfo{person}{Muhammad Ghifary},
  \bibinfo{person}{W~Bastiaan Kleijn}, {and} \bibinfo{person}{Mengjie Zhang}.}
  \bibinfo{year}{2014}\natexlab{}.
\newblock \showarticletitle{Domain adaptive neural networks for object
  recognition}. In \bibinfo{booktitle}{\emph{Pacific Rim International
  Conference on Artificial Intelligence}}. Springer, \bibinfo{pages}{898--904}.
\newblock


\bibitem[\protect\citeauthoryear{Gionis, Indyk, Motwani, et~al\mbox{.}}{Gionis
  et~al\mbox{.}}{1999}]%
        {gionis1999similarity}
\bibfield{author}{\bibinfo{person}{Aristides Gionis}, \bibinfo{person}{Piotr
  Indyk}, \bibinfo{person}{Rajeev Motwani}, {et~al\mbox{.}}}
  \bibinfo{year}{1999}\natexlab{}.
\newblock \showarticletitle{Similarity search in high dimensions via hashing}.
  In \bibinfo{booktitle}{\emph{VLDB}}, Vol.~\bibinfo{volume}{99}.
  \bibinfo{pages}{518--529}.
\newblock


\bibitem[\protect\citeauthoryear{Glorot, Bordes, and Bengio}{Glorot
  et~al\mbox{.}}{2011}]%
        {glorot2011domain}
\bibfield{author}{\bibinfo{person}{Xavier Glorot}, \bibinfo{person}{Antoine
  Bordes}, {and} \bibinfo{person}{Yoshua Bengio}.}
  \bibinfo{year}{2011}\natexlab{}.
\newblock \showarticletitle{Domain adaptation for large-scale sentiment
  classification: A deep learning approach}. In
  \bibinfo{booktitle}{\emph{Proceedings of the 28th international conference on
  machine learning (ICML-11)}}. \bibinfo{pages}{513--520}.
\newblock


\bibitem[\protect\citeauthoryear{Greene and Cunningham}{Greene and
  Cunningham}{2006}]%
        {greene06icml}
\bibfield{author}{\bibinfo{person}{Derek Greene} {and}
  \bibinfo{person}{P\'{a}draig Cunningham}.} \bibinfo{year}{2006}\natexlab{}.
\newblock \showarticletitle{Practical Solutions to the Problem of Diagonal
  Dominance in Kernel Document Clustering}. In \bibinfo{booktitle}{\emph{Proc.
  23rd International Conference on Machine learning (ICML'06)}}.
  \bibinfo{publisher}{ACM Press}, \bibinfo{pages}{377--384}.
\newblock


\bibitem[\protect\citeauthoryear{Greff, Srivastava, Koutn{\'\i}k, Steunebrink,
  and Schmidhuber}{Greff et~al\mbox{.}}{2015}]%
        {greff2015lstm}
\bibfield{author}{\bibinfo{person}{Klaus Greff}, \bibinfo{person}{Rupesh~Kumar
  Srivastava}, \bibinfo{person}{Jan Koutn{\'\i}k}, \bibinfo{person}{Bas~R
  Steunebrink}, {and} \bibinfo{person}{J{\"u}rgen Schmidhuber}.}
  \bibinfo{year}{2015}\natexlab{}.
\newblock \showarticletitle{LSTM: A search space odyssey}.
\newblock \bibinfo{journal}{\emph{arXiv preprint arXiv:1503.04069}}
  (\bibinfo{year}{2015}).
\newblock


\bibitem[\protect\citeauthoryear{Indyk and Motwani}{Indyk and Motwani}{1998}]%
        {indyk1998approximate}
\bibfield{author}{\bibinfo{person}{Piotr Indyk} {and} \bibinfo{person}{Rajeev
  Motwani}.} \bibinfo{year}{1998}\natexlab{}.
\newblock \showarticletitle{Approximate nearest neighbors: towards removing the
  curse of dimensionality}. In \bibinfo{booktitle}{\emph{Proceedings of the
  thirtieth annual ACM symposium on Theory of computing}}. ACM,
  \bibinfo{pages}{604--613}.
\newblock


\bibitem[\protect\citeauthoryear{Kibriya, Frank, Pfahringer, and
  Holmes}{Kibriya et~al\mbox{.}}{2004}]%
        {kibriya2004multinomial}
\bibfield{author}{\bibinfo{person}{Ashraf~M Kibriya}, \bibinfo{person}{Eibe
  Frank}, \bibinfo{person}{Bernhard Pfahringer}, {and}
  \bibinfo{person}{Geoffrey Holmes}.} \bibinfo{year}{2004}\natexlab{}.
\newblock \showarticletitle{Multinomial naive bayes for text categorization
  revisited}. In \bibinfo{booktitle}{\emph{Australasian Joint Conference on
  Artificial Intelligence}}. Springer, \bibinfo{pages}{488--499}.
\newblock


\bibitem[\protect\citeauthoryear{Kingma and Ba}{Kingma and Ba}{2014}]%
        {kingma2014adam}
\bibfield{author}{\bibinfo{person}{Diederik Kingma} {and}
  \bibinfo{person}{Jimmy Ba}.} \bibinfo{year}{2014}\natexlab{}.
\newblock \showarticletitle{Adam: A method for stochastic optimization}.
\newblock \bibinfo{journal}{\emph{arXiv preprint arXiv:1412.6980}}
  (\bibinfo{year}{2014}).
\newblock


\bibitem[\protect\citeauthoryear{Lichman}{Lichman}{2013}]%
        {Lichman:2013}
\bibfield{author}{\bibinfo{person}{M. Lichman}.}
  \bibinfo{year}{2013}\natexlab{}.
\newblock \bibinfo{title}{{UCI} Machine Learning Repository}.
\newblock   (\bibinfo{year}{2013}).
\newblock
\urldef\tempurl%
\url{http://archive.ics.uci.edu/ml}
\showURL{%
\tempurl}


\bibitem[\protect\citeauthoryear{Long, Zhu, Wang, and Jordan}{Long
  et~al\mbox{.}}{2016}]%
        {long2016unsupervised}
\bibfield{author}{\bibinfo{person}{Mingsheng Long}, \bibinfo{person}{Han Zhu},
  \bibinfo{person}{Jianmin Wang}, {and} \bibinfo{person}{Michael~I Jordan}.}
  \bibinfo{year}{2016}\natexlab{}.
\newblock \showarticletitle{Unsupervised domain adaptation with residual
  transfer networks}. In \bibinfo{booktitle}{\emph{Advances in Neural
  Information Processing Systems}}. \bibinfo{pages}{136--144}.
\newblock


\bibitem[\protect\citeauthoryear{Maaten and Hinton}{Maaten and Hinton}{2008}]%
        {maaten2008visualizing}
\bibfield{author}{\bibinfo{person}{Laurens van~der Maaten} {and}
  \bibinfo{person}{Geoffrey Hinton}.} \bibinfo{year}{2008}\natexlab{}.
\newblock \showarticletitle{Visualizing data using t-SNE}.
\newblock \bibinfo{journal}{\emph{Journal of Machine Learning Research}}
  \bibinfo{volume}{9}, \bibinfo{number}{Nov} (\bibinfo{year}{2008}),
  \bibinfo{pages}{2579--2605}.
\newblock


\bibitem[\protect\citeauthoryear{McGaugh}{McGaugh}{2000}]%
        {mcgaugh2000memory}
\bibfield{author}{\bibinfo{person}{James~L McGaugh}.}
  \bibinfo{year}{2000}\natexlab{}.
\newblock \showarticletitle{Memory--a century of consolidation}.
\newblock \bibinfo{journal}{\emph{Science}} \bibinfo{volume}{287},
  \bibinfo{number}{5451} (\bibinfo{year}{2000}), \bibinfo{pages}{248--251}.
\newblock


\bibitem[\protect\citeauthoryear{Pan and Yang}{Pan and Yang}{2010}]%
        {pan2010survey}
\bibfield{author}{\bibinfo{person}{Sinno~Jialin Pan} {and}
  \bibinfo{person}{Qiang Yang}.} \bibinfo{year}{2010}\natexlab{}.
\newblock \showarticletitle{A survey on transfer learning}.
\newblock \bibinfo{journal}{\emph{IEEE Transactions on knowledge and data
  engineering}} \bibinfo{volume}{22}, \bibinfo{number}{10}
  (\bibinfo{year}{2010}), \bibinfo{pages}{1345--1359}.
\newblock


\bibitem[\protect\citeauthoryear{Pedregosa, Varoquaux, Gramfort, Michel,
  Thirion, Grisel, Blondel, Prettenhofer, Weiss, Dubourg, Vanderplas, Passos,
  Cournapeau, Brucher, Perrot, and Duchesnay}{Pedregosa et~al\mbox{.}}{2011}]%
        {scikit-learn}
\bibfield{author}{\bibinfo{person}{F. Pedregosa}, \bibinfo{person}{G.
  Varoquaux}, \bibinfo{person}{A. Gramfort}, \bibinfo{person}{V. Michel},
  \bibinfo{person}{B. Thirion}, \bibinfo{person}{O. Grisel},
  \bibinfo{person}{M. Blondel}, \bibinfo{person}{P. Prettenhofer},
  \bibinfo{person}{R. Weiss}, \bibinfo{person}{V. Dubourg}, \bibinfo{person}{J.
  Vanderplas}, \bibinfo{person}{A. Passos}, \bibinfo{person}{D. Cournapeau},
  \bibinfo{person}{M. Brucher}, \bibinfo{person}{M. Perrot}, {and}
  \bibinfo{person}{E. Duchesnay}.} \bibinfo{year}{2011}\natexlab{}.
\newblock \showarticletitle{Scikit-learn: Machine Learning in {P}ython}.
\newblock \bibinfo{journal}{\emph{Journal of Machine Learning Research}}
  \bibinfo{volume}{12} (\bibinfo{year}{2011}), \bibinfo{pages}{2825--2830}.
\newblock


\bibitem[\protect\citeauthoryear{Raina, Battle, Lee, Packer, and Ng}{Raina
  et~al\mbox{.}}{2007}]%
        {raina2007self}
\bibfield{author}{\bibinfo{person}{Rajat Raina}, \bibinfo{person}{Alexis
  Battle}, \bibinfo{person}{Honglak Lee}, \bibinfo{person}{Benjamin Packer},
  {and} \bibinfo{person}{Andrew~Y Ng}.} \bibinfo{year}{2007}\natexlab{}.
\newblock \showarticletitle{Self-taught learning: transfer learning from
  unlabeled data}. In \bibinfo{booktitle}{\emph{Proceedings of the 24th
  international conference on Machine learning}}. ACM,
  \bibinfo{pages}{759--766}.
\newblock


\bibitem[\protect\citeauthoryear{Rosenstein, Marx, Kaelbling, and
  Dietterich}{Rosenstein et~al\mbox{.}}{2005}]%
        {rosenstein2005transfer}
\bibfield{author}{\bibinfo{person}{Michael~T Rosenstein},
  \bibinfo{person}{Zvika Marx}, \bibinfo{person}{Leslie~Pack Kaelbling}, {and}
  \bibinfo{person}{Thomas~G Dietterich}.} \bibinfo{year}{2005}\natexlab{}.
\newblock \showarticletitle{To transfer or not to transfer}. In
  \bibinfo{booktitle}{\emph{NIPS 2005 Workshop on Transfer Learning}},
  Vol.~\bibinfo{volume}{898}.
\newblock


\bibitem[\protect\citeauthoryear{Shindler, Wong, and Meyerson}{Shindler
  et~al\mbox{.}}{2011}]%
        {shindler2011fast}
\bibfield{author}{\bibinfo{person}{Michael Shindler}, \bibinfo{person}{Alex
  Wong}, {and} \bibinfo{person}{Adam~W Meyerson}.}
  \bibinfo{year}{2011}\natexlab{}.
\newblock \showarticletitle{Fast and accurate k-means for large datasets}. In
  \bibinfo{booktitle}{\emph{Advances in neural information processing
  systems}}. \bibinfo{pages}{2375--2383}.
\newblock


\bibitem[\protect\citeauthoryear{Suykens and Vandewalle}{Suykens and
  Vandewalle}{1999}]%
        {suykens1999least}
\bibfield{author}{\bibinfo{person}{Johan~AK Suykens} {and}
  \bibinfo{person}{Joos Vandewalle}.} \bibinfo{year}{1999}\natexlab{}.
\newblock \showarticletitle{Least squares support vector machine classifiers}.
\newblock \bibinfo{journal}{\emph{Neural processing letters}}
  \bibinfo{volume}{9}, \bibinfo{number}{3} (\bibinfo{year}{1999}),
  \bibinfo{pages}{293--300}.
\newblock


\bibitem[\protect\citeauthoryear{Wang, Du, Zhang, Zhang, Hu, and Tao}{Wang
  et~al\mbox{.}}{[n. d.]}]%
        {wanggleaning}
\bibfield{author}{\bibinfo{person}{Zengmao Wang}, \bibinfo{person}{Bo Du},
  \bibinfo{person}{Lefei Zhang}, \bibinfo{person}{Liangpei Zhang},
  \bibinfo{person}{Ruimin Hu}, {and} \bibinfo{person}{Dacheng Tao}.}
  \bibinfo{year}{[n. d.]}\natexlab{}.
\newblock \showarticletitle{On Gleaning Knowledge from Multiple Domains for
  Active Learning}.
\newblock  (\bibinfo{year}{[n. d.]}).
\newblock


\bibitem[\protect\citeauthoryear{Wang, Song, and Zhang}{Wang
  et~al\mbox{.}}{2008}]%
        {wang2008transferred}
\bibfield{author}{\bibinfo{person}{Zheng Wang}, \bibinfo{person}{Yangqiu Song},
  {and} \bibinfo{person}{Changshui Zhang}.} \bibinfo{year}{2008}\natexlab{}.
\newblock \showarticletitle{Transferred dimensionality reduction}. In
  \bibinfo{booktitle}{\emph{Joint European Conference on Machine Learning and
  Knowledge Discovery in Databases}}. Springer, \bibinfo{pages}{550--565}.
\newblock


\end{thebibliography}

\end{document}